\documentclass{article}

\usepackage[nonatbib,preprint]{neurips_data_2022}

\usepackage[utf8]{inputenc} 
\usepackage[T1]{fontenc}    
\usepackage{hyperref}       
\usepackage{url}            
\usepackage{booktabs}       
\usepackage{amsfonts}       
\usepackage{nicefrac}       
\usepackage{microtype}      
\usepackage{xcolor}         

\usepackage{graphicx} 
\usepackage{float} 
\usepackage{subfigure} 
\usepackage{color}
\usepackage{threeparttable}
\usepackage{booktabs}
\usepackage{multirow}
\usepackage{amsmath,mathtools,amssymb}
\usepackage{subcaption} 
\usepackage{makecell}

\newcommand{\ourproj}[1]{\emph{DropletVideo}}
\newcommand{\ourdataset}[1]{\emph{DropletVideo-10M}}

\usepackage{tabularx}

\title{DropletVideo: A Dataset and Approach to Explore Integral Spatio-Temporal Consistent Video Generation}

\author{
Runze Zhang$^{1,}$\footnotemark[1] ~,
Guoguang Du$^{1,}$\footnotemark[1] ~,
Xiaochuan Li$^{1,}$\footnotemark[1] ~,  
Qi Jia$^{1,}$\footnotemark[1] ~, 
Liang Jin$^{1,}$\footnotemark[1]
\\
\textbf{Lu Liu$^{1}$, Jingjing Wang$^{1}$, Cong Xu$^{1}$, 
Zhenhua Guo$^{1}$, Yaqian Zhao$^{1}$, Xiaoli Gong$^{2}$}
\\
\textbf{Rengang Li$^{1,3,}$\footnotemark[2] ~, Baoyu Fan$^{1,2,}$\footnotemark[2]}
\\
  $^1$ IEIT System Co., Ltd. ~~
  $^2$ Nankai University ~~
  $^3$ Tsinghua University\\
  \url{https://dropletx.github.io}
}

\begin{document}

\maketitle
\renewcommand{\thefootnote}{\fnsymbol{footnote}}
\footnotetext[1]{Equal contribution.}
\footnotetext[2]{Corresponding author.}

\begin{abstract}

Spatio-temporal consistency is a critical research topic in video generation. 
A qualified generated video segment must ensure plot plausibility and coherence while maintaining visual consistency of objects and scenes across varying viewpoints.
Prior research, especially in open-source projects, primarily focuses on either temporal or spatial consistency, or their basic combination, such as appending a description of a camera movement after a prompt without constraining the outcomes of this movement.
However, camera movement may introduce new objects to the scene or eliminate existing ones, thereby overlaying and affecting the preceding narrative. 
Especially in videos with numerous camera movements, the interplay between multiple plots becomes increasingly complex. 
This paper introduces and examines integral spatio-temporal consistency, considering the synergy between plot progression and camera techniques, and the long-term impact of prior content on subsequent generation.
Our research encompasses dataset construction through to the development of the model. 
Initially, we constructed a DropletVideo-10M dataset, which comprises 10 million videos featuring dynamic camera motion and object actions. 
Each video is annotated with an average caption of 206 words, detailing various camera movements and plot developments. 
Following this, we developed and trained the DropletVideo model, which excels in preserving spatio-temporal coherence during video generation. 
The DropletVideo dataset and model are accessible at \url{https://dropletx.github.io}.

\end{abstract}

\section{Introduction}

\begin{figure}[ht]
 \centering
 \includegraphics[width=1.0\linewidth]{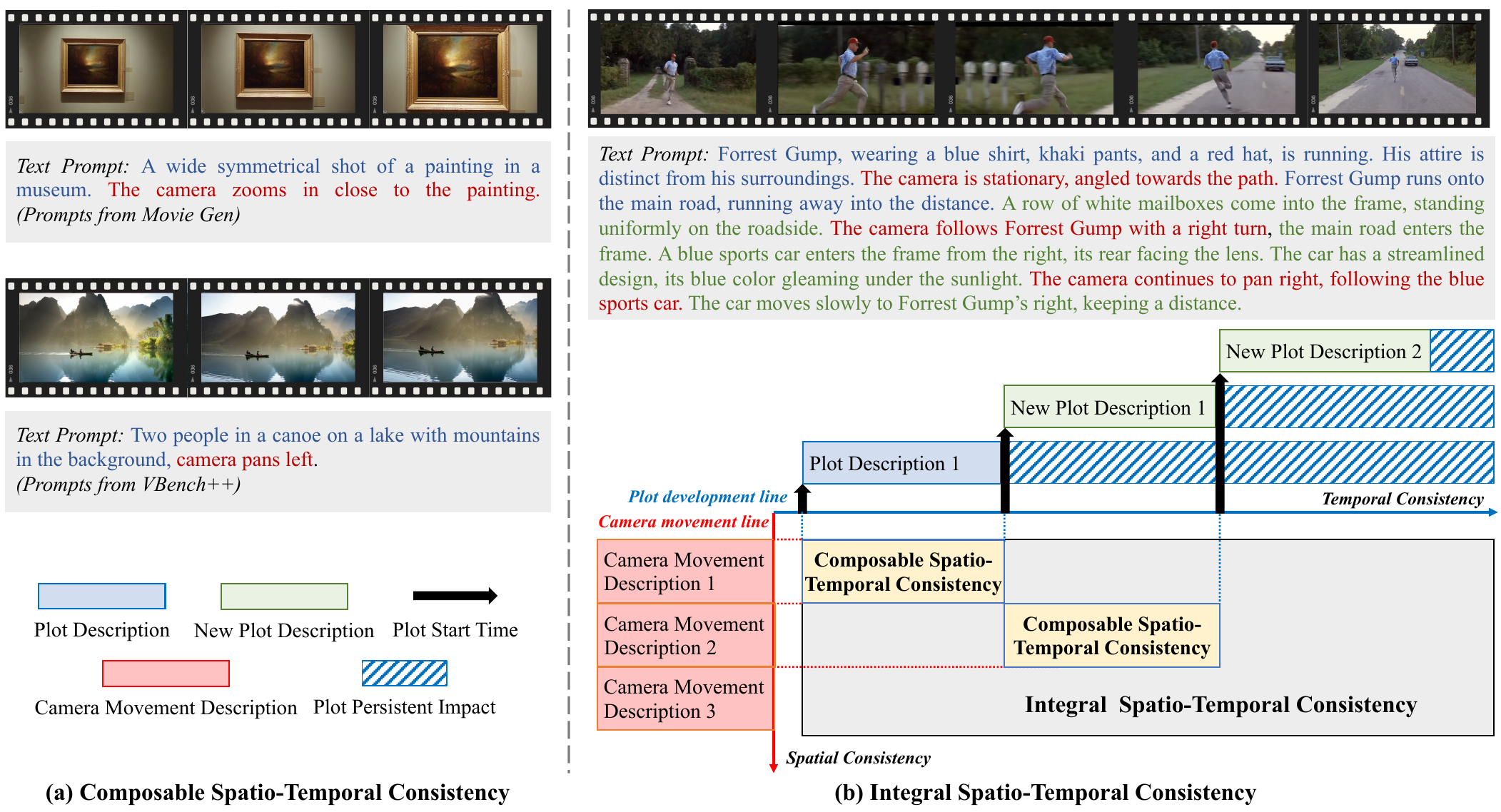}
 \caption{\textbf{Comparisons between Composable Spatio-temporal Consistency and Integral Spatio-temporal Consistency.} 
 (a) \textbf{Composable Spatio-Temporal Consistency} refers to the straightforward combination of temporal and spatial consistency, without limiting the effects of camera movement. Studies such as MovieGen \cite{polyak2024moviegencastmedia} and VBench++ \cite{huang2024vbench++} are dedicated to realizing this consistency. Despite the potential emergence of a new scene post camera movement, the introduced scene tends to be stationary, precluding the onset of further motion.
 (b) \textbf{Integral Spatio-Temporal Consistency} considers the interplay between plot development and camera techniques, along with the enduring influence of antecedent content on subsequent creation. 
 This is because a camera movement may introduce or eliminate objects, thereby overlaying and impacting the preceding storyline.
 For example in the ``Forrest Gump'' clip, achieving integral spatio-temporal consistency requires incorporating the motion of the ``car'' as it recedes following the camera’s ``turn right'' action while maintaining the scene of Forrest running, ensuring that ``Forrest Gump's right remains at a consistent distance'', preserving the correct spatial relationships.
 Temporal consistency in plot progression is highlighted in the \textcolor{blue}{blue} region, while the \textcolor{red}{red} region denotes spatial consistency induced by camera movement}
 \label{fig_intergralsc}
\end{figure}

Video generation is a crucial task in AI-generated content (AIGC). 
Unlike static image generation, video generation involves dynamic variations across frames, making it significantly more complex. 
The primary challenge lies in maintaining spatio-temporal consistency, ensuring both spatial coherence within each frame and temporal continuity across consecutive frames. 
This challenge can be further decomposed into two key aspects:

\noindent{\textbf{Temporal Consistency}}: Ensuring smooth transitions between frames that adhere to physical principles, enabling the video to progress in a plausible and consistent manner. 
This is exemplified by the consistent depiction of Forest’s actions in the blue region of Fig. \ref{fig_intergralsc} (b).

\noindent{\textbf{Spatial Consistency}}: Maintaining consistent visual characteristics of objects and scenes (e.g., shape, size, texture, and color) across different viewpoints is essential for spatio-temporal coherence. This is demonstrated in the red region of Fig. \ref{fig_intergralsc} (b), where camera rotation or upward movement preserves object consistency.

\begin{figure}[h!]
    \centering
    \includegraphics[width=0.98\linewidth]{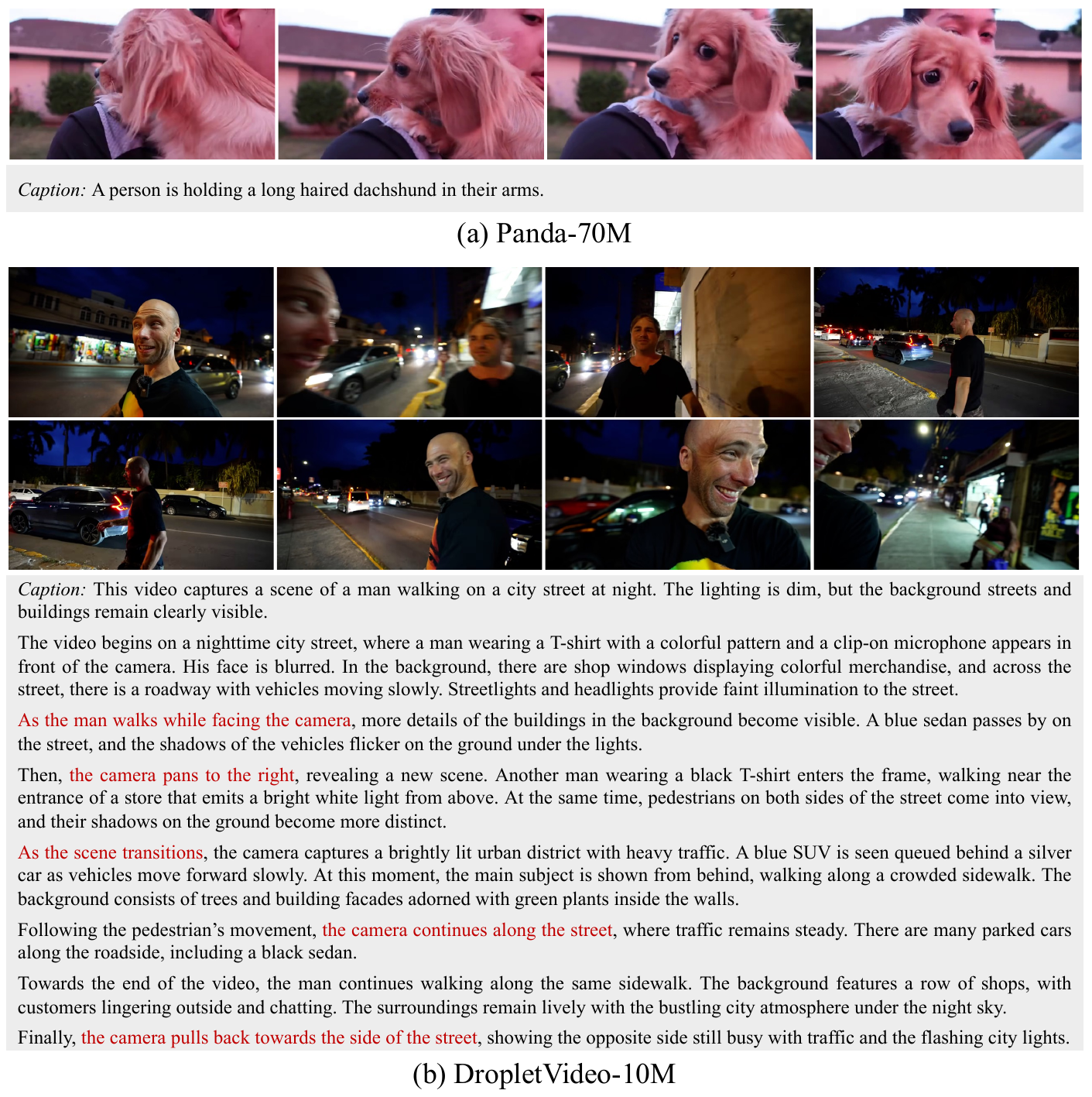}
    \caption{\textbf{The \ourdataset{} dataset features diverse camera movements, long-captioned contextual descriptions, and strong spatio-temporal consistency.} (a) Existing datasets, such as Panda-70M \cite{chen2024panda-70m}, place less emphasis on camera movement and contain relatively brief captions. (b) In contrast, \ourdataset{} consists of spatio-temporal videos that incorporate both camera movement and event progression. Each video is paired with a caption that conveys detailed spatio-temporal information aligned with the video content, with an average caption length of 206 words. The spatio-temporal information is highlighted in red in the figure.}
    \label{fig2}
\end{figure}

Studies in video generation have increasingly focused on addressing the challenge of visual transition consistency. 
Blattmann et al. \cite{Blattmann2023AlignYL} and Luo et al. \cite{Luo2023VideoFusionDD} have contributed to enhancing image quality and improving event transition plausibility, ensuring greater narrative coherence and accuracy.
Cheong et al. \cite{cheong2024boosting} and Wang et al. \cite{wang2024cpa} have explored the unification of objects across different viewpoints while accounting for camera movement.
Commercial models such as Sora \cite{openai-sora} and Kling-1.5 \cite{kuaishou-klingai} demonstrate strong spatio-temporal consistency. However, as these models are closed-source, they restrict public access and limit algorithmic innovation \cite{kong2024hunyuanvideo}.

Recently, research has underscored the importance of spatio-temporal consistency, with a focus on effecting camera angle shifts concurrent with plot progression, as delineated by the light yellow box in Fig. (b). 
Nonetheless, these inquiries are confined to the realm of composable spatio-temporal consistency, namely appending camera movement descriptions to prompts without circumscribing the movement's consequences.
The current benchmark for video generation evaluation, VBench++ \cite{huang2024vbench++}, includes an assessment of this capability.

Howerver, a camera movement has the potential to introduce new elements or remove existing ones from the scene, thereby altering the preceding narrative. 
For example, ``Forrest’s uninterrupted running'' while a car enters and exits the frame due to a camera turn, as illustrated in Fig. \ref{fig_intergralsc} (b). 
Besides, in videos featuring numerous camera movements, the interaction among various plot elements becomes increasingly intricate.
Therefore, we introduce and explore integral spatio-temporal consistency, focusing on the synergy between plot progression and camera techniques, and the enduring impact of earlier content on subsequent generation.
From the perspective of video generation tasks, ensuring this consistency has the potential to advance generated content from single-shot, plot-based videos to more complex, multi-plot narratives.

Our research on this issue encompasses from dataset construction to model development.
We propose an open-source integral spatio-temporal consistency video dataset, named \ourdataset{}. 
To the best of our knowledge, it is the largest open-source dataset that preserves integral spatio-temporal consistency.
A key attribute of \ourdataset{} is its inclusion of videos featuring both object motion and camera movement. 
Compared to traditional datasets that primarily contain videos with object motion alone, such dual-motion video samples are underrepresented in existing datasets. 
Additionally, to support the training of integral spatio-temporal consistency models, captions must provide meticulously detailed information, including both object motion and camera movement. 
Traditional video captions often omit such specifics, typically focusing on scenery and plot while failing to capture the nuances of motion, particularly those induced by camera movement. 
\ourdataset{} addresses this limitation by providing captions that explicitly describe these motion aspects, including the effects of camera movements.
With an average caption length of 206 words, \ourdataset{} surpasses existing datasets in descriptive depth. 
\ourdataset{} offers an extensive collection of videos encompassing both motion types, thereby providing a more balanced and comprehensive dataset for video generation research.

Based on \ourdataset{}, we propose a pre-trained model, \ourproj{}, an open-source foundational model for video generation. 
It is designed to maintain integral spatio-temporal consistency while simultaneously generating both camera movement and plot progression.
\ourproj{} also incorporates a variable frame rate sampling strategy, enabling precise control over video generation speed and the tempo of visual transitions.
Comprehensive experiments have been conducted, and the results confirm that \ourproj{} effectively preserves content consistency across both temporal and spatial dimensions. 

The contributions of this work are as follows:
\begin{itemize}
    \item We introduce \textit{Integral Spatio-Temporal Consistency} in video generation, an aspect that has not been previously explored. By emphasizing Integral Spatio-Temporal Consistency, we enable the generation of more complex, multi-plot narratives with natural camera movements and smooth scene transitions.
    \item We have constructed \ourdataset{}, the largest dataset designed for integral spatio-temporal consistency in video generation. It is 43$\times$ larger than MVImageNet~\cite{yu2023mvimgnet} and comparable in scale to the large-scale video generation dataset Panda-70M~\cite{chen2024panda-70m}. Additionally, our dataset features an average caption length of 206 words, which is 15.6$\times$ longer than that of Panda-70M, providing significantly richer textual descriptions.
    \item We propose \ourproj{}, a pre-trained foundational video generation model based on \ourdataset{}, which excels in producing videos with integral spatio-temporal consistency.
    \item We have open-sourced the dataset, code, and model weights of \ourproj{}. We hope this initiative fosters algorithmic innovation in the public domain, encouraging further advancements that match or even surpass closed-source models. 
\end{itemize}


\section{Related Work}

\subsection{Video-Language Datasets}

To advance video generation tasks, especially text-conditioned video generation, several video-language datasets have been introduced in recent years\cite{miech2019howto100m, bain2021frozen-Webvid-10m, xue2022advancing-HD-VILA-100M, wang2023internvid, wang2023videofactory-HD-VG-130M, chen2024panda-70m, ju2024miradata}.
For instance, Panda-70M\cite{chen2024panda-70m} presents a large-scale dataset with 70 million video clips annotated with automatic captions. 
This dataset covers 166.8Khr with average 13.2 words.
MiraData\cite{ju2024miradata}, on the other hand, offers a high-quality dataset comprising 788K videos, each accompanied by detailed captions, averaging 318 words per caption.
These works have significantly advanced the field of video generation. 
Nonetheless, they primarily focus on temporal consistency in videos, overlooking data subject to perspective transformations.

On the other hand, to tackle the spatial consistency challenge in video generation, several multi-view image datasets\cite{reizenstein2021common-CO3D, yu2023mvimgnet} and video datasets\cite{ling2024dl3dv, zhou2018stereo-RealEstate10K} have been proposed. 
CO3Dv2\cite{reizenstein2021common-CO3D} and MVImageNet\cite{yu2023mvimgnet} predominantly feature object-level multi-view images, while DL3DV-10K\cite{ling2024dl3dv} and Real-estate-10K\cite{zhou2018stereo-RealEstate10K} focused on scene-level videos. 
However, the majority of these datasets contain a restricted number of video frames and are predominantly utilized for multi-view image generation, rather than video generation. 
As a result, the objects within the scenes are stationary, disregarding temporal consistency. 
Moreover, these datasets are substantially smaller in volume compared to those specifically curated to address temporal consistency challenges.

Recently, an increasing number of researchers have focused on tackling the spatio-temporal consistency problem in video generation, with several datasets being introduced. 
For example, MV-Video\cite{jiang2024animate3d-MV-Video} comprises around 115K publicly available animations, including about 53K animated 3D objects, rendered into over 1.8 million multi-view videos.
These efforts, which consider both object and camera movement, have contributed to the advancement of video generation.
However, they neglect the description of the interplay between the two, such as the cumulative plot changes resulting from camera movement.

In comparison, we have curated the world’s largest video-language dataset, \ourdataset{}, as shown in Tab. \ref{tab:dataset_statistics}, which addresses the spatio-temporal consistency problem integrally. 
All videos in \ourdataset{} involve camera motions and the quantity is 43$\times$ larger than the multi-view images dataset MVImageNet \cite{yu2023mvimgnet}, and comparable with the large-scale video generation dataset Panda-70M \cite{chen2024panda-70m}. 
Additionly, the average caption of our dataset is 206 words, which is 15.6$\times$ in comparison with Panda-70M \cite{chen2024panda-70m}.

\begin{table}[ht]
\centering
\caption{\textbf{Comparison of \ourdataset{} and other video-language datasets.} \ourdataset{} dataset possesses unique advantages. 
\textbf{First}, it contains longer text captions than all but MiraData, yet MiraData is substantially smaller in scale. 
\textbf{Second}, with an average video length of 7.3 seconds, it exhibits the highest information density per second of video. 
\textbf{Third}, \ourdataset{} emphasizes the spatio-temporal attributes of videos and captions, distinguishing it as the most comprehensive spatio-temporal video generation dataset to date. In contrast, datasets like Koala-36M, despite their wealth of textual descriptions, do not prioritize the specifics of spatial transformations due to camera movement.
}
\label{tab:dataset_statistics}
\small
\begin{tabular}{lccccccc}
\toprule
  & Words & Year & Clips & Avg dur. & Total dur. & Category \\
\midrule
HowTo100M~\cite{miech2019howto100m} & 4.0 words & 2019 & 100M & 3.6s & 135Khr & Temporal\\
WebVid-10M~\cite{bain2021frozen-Webvid-10m} & 12.0 words & 2021 & 10M & 18.0s & 52Khr & Temporal\\
HD-VILA-100M~\cite{xue2022advancing-HD-VILA-100M} & 17.6 words & 2022 & 100M & 11.7s & 760.3Khr & Temporal\\
InternVid~\cite{wang2023internvid} & 32.5 words & 2023 & 7M & 13.4s & 371.5Khr & Temporal\\
HD-VG-130M~\cite{wang2023videofactory-HD-VG-130M} & \raisebox{0.5ex}{\texttildelow}9.6 words & 2024 & 130M & \raisebox{0.5ex}{\texttildelow}5.1s & \raisebox{0.5ex}{\texttildelow}184Khr & Temporal\\
Panda-70M~\cite{chen2024panda-70m} & 13.2 words & 2024 & 70M & 8.5s & 167Khr & Temporal\\
MiraData~\cite{ju2024miradata} & 318.0 words & 2024 & 788K & 72.1s & 16Khr & Temporal\\
Koala-36M~\cite{wang2024koala} & 202.1 words & 2024 & 36M & 13.75s & 172Khr & Temporal\\
\midrule
CO3Dv2~\cite{reizenstein2021common-CO3D} & - & 2021 & 36k & - & - & Spatial\\
DL3DV-10K~\cite{ling2024dl3dv} & - & 2023 & 10K & - & - & Spatial\\
RealEstate-10K~\cite{zhou2018stereo-RealEstate10K} & -& 2023 & 10K & - & - & Spatial\\
MVImageNet~\cite{yu2023mvimgnet} & - & 2023 & 229K & - & - & Spatial\\
\midrule
MV-Video~\cite{jiang2024animate3d-MV-Video} & - & 2024 & 1.8M & 2s & 1Khr & Spatio-Temporal\\
\textbf{DropletVideo-10M (Ours)} & 206.0 words & 2025 & 10M & 7.3s & 20.4Khr & Spatio-Temporal\\
\bottomrule
\end{tabular}
\end{table}

\subsection{Spatio-temporal Consistent Video Generation}
Due to the high continuity and dynamic variability of video data, directly generating dynamically consistent videos in both temporal and spatial dimensions is a highly challenging task. 
As a result, generated videos often fail to meet practical requirements.

Many video generation studies primarily focus on temporal consistency. 
Blattmann et al.\cite{Blattmann2023AlignYL} propose a high-resolution video framework utilizing pre-trained Latent Diffusion Models (LDM). 
This framework introduces a temporal dimension to the latent space and incorporates learnable temporal layers, ensuring inter-frame alignment.
Videofusion\cite{Luo2023VideoFusionDD} introduces a decomposed diffusion model, which separates spatial and temporal optimizations to improve cross-frame consistency. 
It leverages time-aware latent representations and a hierarchical strategy, effectively minimizing temporal jitter.

For spatial consistency, researchers have initially proposed a series of models based on the U-Net architecture~\cite{liu2022video, he2024cameractrl, wang2024objctrl, yu2024viewcrafter}.
Diffusion Transformer (DiT)\cite{hatamizadeh2024diffit} combines the benefits of Visual Transformer (ViT) and Diffusion Diffusion Model (DDPM), gradually replacing U-Net as the predominant architecture in video generation tasks.
Cheong et al.\cite{cheong2024boosting} address the low motion accuracy of DiT by introducing camera motion guidance and a sparse camera control pipeline.
DiT-based video generation methods have made substantial progress in generating high-quality long videos\cite{bahmani2024ac3d, wang2024cpa, bahmani2024vd3d}.

Naturally, jointly considering spatio-temporal consistency has become a critical challenge in generation tasks. In 4D generation tasks, spatio-temporal consistency remains a central and unavoidable research focus\cite{jiang2024consistentd, pan2024fast, liang2024diffusion4d, yang2024diffusion, yin20234dgen, zeng2024stag4d}. 
Recent advances in video generation research have similarly concentrated on spatial-temporal consistency.
Singer et al.\cite{Singer2022MakeAVideoTG} leverage pretrained text-to-image diffusion models and introduce pseudo-3D convolutional layers to enhance temporal coherence without requiring text-video paired data.
ModelScope\cite{Wang2023ModelScopeTT} proposes a hybrid architecture that combines spatial-temporal blocks with cross-frame attention to maintain multi-scale consistency.
Qing et al.\cite{Qing2023HierarchicalSD} propose a two-stage framework that explicitly disentangles spatial and temporal modeling, first generating keyframes and then interpolating the motion between them.
Agrim et al.\cite{10.1007/978-3-031-72986-7_23} improve temporal alignment using a cascaded diffusion pipeline with optical flow-guided latent propagation.
Chen et al.\cite{chen2024unictrl} address spatio-temporal inconsistencies with a training-free approach that integrates spatial and temporal attention controls during diffusion sampling.

However, these methods address either temporal or spatial consistency, or their rudimentary combination, as in maintaining consistent camera movement during storyline generation. 
Nonetheless, these approaches are inadequate for complex captions. 
In the work, we attempt to design and train a novel model to ensure the generation of videos with integral spatio-temporal consistency.

\subsection{Open-Source Landscape of Video Generation Models}

Although large amounts of video generation models\cite{kong2024hunyuanvideo,kuaishou-klingai,lumalabs.ai-dream-machine,moviegen,runway2024gen} are proposed, most of them are commercial closed-source models.
Kling v1.6\cite{kuaishou-klingai} focuses on personalized 10 seconds video creation, leveraging user data to offer tailored templates and effects, and enabling easy creation through gesture and voice commands.
Luma Dream Machine\cite{lumalabs.ai-dream-machine} combines deep learning and reinforcement learning to generate videos that reflect user emotions and intentions.
Meta's Movie Gen\cite{moviegen} explores the potential of text-to-video synthesis with a focus on scalability and accessibility.
Runway Gen-3 \cite{runway2024gen} allows users to generate, edit, and transform videos using simple text-based instructions, bridging the gap between technical algorithms and creative workflows.

Except from these commercial models, some of the video generation models in community are open-sourced.
However, their performance lags significantly behind commercial models. 
Moreover, few of them totally open-source their models and training data, as shown in Tab. \ref{tab:open_source_summary}.
Whereas, we open-source all the information about \ourproj{}, hoping to raise the development of video generation technology and open up new possibilities for research and application in the field.

\begin{table}[ht]
\centering
\caption{\textbf{Open-Source Landscape of Video Generation Models.} We have fully open-sourced the model, technological solution, and data, making it, to the best of our knowledge, the video generation solution with the highest degree of open-source accessibility available. Notably, Our dataset is \textbf{self-collected} and has not previously appeared in the community.}
\label{tab:open_source_summary}
\small
\begin{tabular}{lcccccc}
\toprule
 & Institute & Year & Model & Tech Solution & Data & Self-Collected Data \\
\midrule
I2VGen-XL~\cite{moviegen} & Alibaba & 2023 & $\checkmark$ & $\checkmark$ & $\times$ & $ \times$\\
Animate-Anything~\cite{lei2024animateanything} & Alibaba & 2024 & $\checkmark$ & $\checkmark$ & $\times$ & $ \times$\\
SVD-XT-1.1~\cite{blattmann2023stable} & Stability AI & 2024 & $\checkmark$ & $\checkmark$ & $\times$ & $\times$\\
DynamiCrafter~\cite{xing2024dynamicrafter} & Tencent & 2024 & $\checkmark$ & $\checkmark$ & $\checkmark$ & $\times$\\
CogVideoX~\cite{yang2024cogvideox} & Zhipu AI & 2024 & $\checkmark$ & $\checkmark$ & $\times$ & $\times$\\
HunyuanVideo~\cite{kong2024hunyuanvideo} & Tencent & 2024 & $\checkmark$ & $\checkmark$ & $\times$ & $\times$\\
OpenSora~\cite{2024opensora} & HPC-AI Tech & 2024 & $\checkmark$ & $\checkmark$ & $\checkmark$ & $\times$\\
OpenSoraPlan~\cite{2024Open-sora-plan} & PKU & 2024 & $\checkmark$ & $\checkmark$ & $\checkmark$ & $\times$\\
WanX \cite{wanx} & Alibaba & 2024 & $\checkmark$ & $\checkmark$ & $\times$ & $\times$\\
Cosmos~\cite{agarwal2025cosmos} & Nvidia & 2025 & $\checkmark$ & $\checkmark$ & $\times$ & $\times$\\
Step-Video~\cite{ma2025stepvideot2v} & Stepfun & 2025 & $\checkmark$ & $\checkmark$ & $\times$ & $\times$\\

\midrule
Movie Gen~\cite{Singer2022MakeAVideoTG} & Meta & 2024 & $\times$ & $\checkmark$ & $\times$ & $\checkmark$\\
Gen-3 ~\cite{runway2024gen} & Runway & 2024 & $\times$ & $\times$ & $\times$ & $-$\\
Sora~\cite{openai-sora} & OpenAI & 2024 & $\times$ & $\times$ & $\times$ & $-$\\
Pika~\cite{pika} & Pika & 2024 & $\times$ & $\times$ & $\times$ & $-$ \\
Vivago~\cite{vivago} & Vivago & 2024 & $\times$ & $\times$ & $\times$ & $-$ \\
Ray2~\cite{lumalabs.ai-dream-machine}& Luma AI & 2025 & $\times$ & $\times$ & $\times$ & $-$\\
Kling~\cite{kuaishou-klingai} & Kwai & 2024 & $\times$ & $\times$ & $\times$ & $-$\\
Vidu~\cite{vidu} & Vidu & 2024 & $\times$ & $\times$ & $\times$ & $-$\\
Hailuo~\cite{hailuo} & MiniMax & 2024 & $\times$ & $\times$ & $\times$ & $-$\\
Qingying \cite{qingying} & Zhipu AI & 2024 & $\times$ & $\times$ & $\times$ & $-$\\

\midrule
\textbf{DropletVideo (Ours)} & & 2025 & $\checkmark$ & $\checkmark$ & $\checkmark$ & $\checkmark$\\
\bottomrule
\end{tabular}
\end{table}


\section{Dataset}

A large proportion of videos in existing video generation datasets, such as OpenVid-1M \cite{nan2024openvid-1m}, Open-Sora-Plan \cite{2024Open-sora-plan}, and Panda-70M \cite{chen2024panda-70m}, primarily focus on object movements within frames while lacking camera motions.
We first filtered approximately 600K high-quality spatio-temporal video clips from OpenVid-1M \cite{nan2024openvid-1m}, MiraData\cite{ju2024miradata}, and Pexels \cite{pexels}.
However, this amount of data is insufficient to train a foundation video generation model. Consequently, we construct a dataset from scratch which incorporates both object movement and dynamic camera viewpoint changes.
Furthermore, existing video captions, serving as textual labels and metadata, often fail to account for spatio-temporal consistency.
We address this limitation by enhancing caption quality in our dataset, ensuring a more comprehensive representation of motion dynamics.

\begin{figure}[h]
    \centering
    \includegraphics[width=1.0\linewidth]{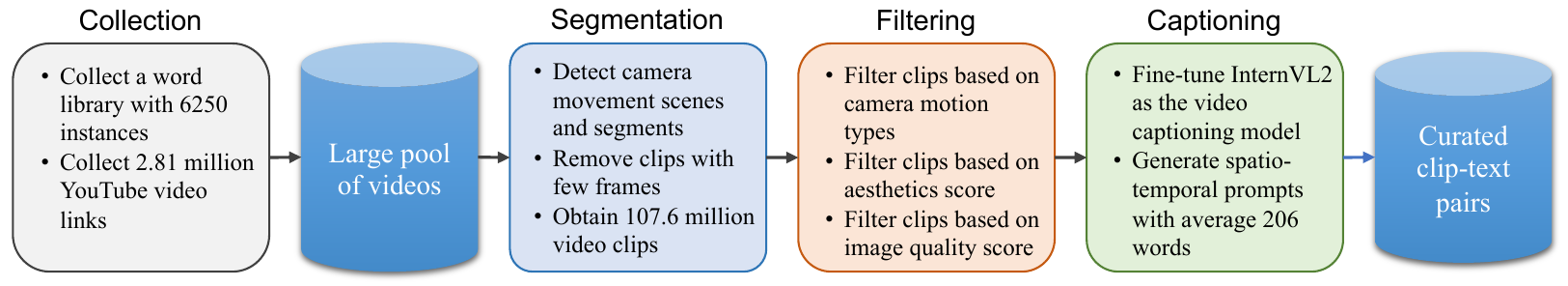}
    \caption{The pipeline we proposed to curate the \ourdataset{} dataset.}
    \label{fig:dataset_pipeline}
\end{figure}

To ensure that the videos in our dataset are both realistic and practical, we construct the dataset using existing video sources, including movies, short films, VLOGs, and similar content.
However, these videos are typically complex, often comprising multiple scenes, which makes them impractical for current video generation tasks. 
To address this, we segment the videos and selectively retain those that properly for video generation training.
Specially, we focus on the videos feature both object motion and camera movement.
To accomplish this task, we propose a dataset curation pipeline, as illustrated in Fig. \ref{fig:dataset_pipeline}. 
This pipeline consists of four key stages: video collection, video segmentation, spatio-temporal variation filtering, and the generation of spatio-temporal consistent captions.


Through our pipeline, we ultimately curated a spatio-temporal dataset containing 10 million high-quality videos, spanning 2.21 billion frames with a total video length of 20.4K hours. 
We name it is as \ourdataset{}.
We have open-sourced the \ourdataset{} dataset to facilitate the research on spatio-temporal consistent video generation. 
Please note that since the original videos in \ourdataset{} were sourced from the internet, they are available exclusively for academic and non-commercial use under the CC BY-NC-SA 4.0 license.

\subsection{Raw Video Collection}
We select YouTube as our primary video source due to its status as one of the largest content platforms, offering a diverse range of videos, including self-recorded footage, aerial shots, animations, gaming content, and more.
To collect videos with spatio-temporal variations, we utilize YouTube’s search functionality. 
We construct a set of \textbf{6,250} search keywords.
We then collect \textbf{2.81 million} video links from the search result, an average of \textbf{450} links per search term.
For comparison, Panda-70M~\cite{chen2024panda-70m}, derived from HD-VILA-100M~\cite{xue2022advancing-HD-VILA-100M}, contains approximately \textbf{3.3 million} video links. 
The scale of our dataset is therefore comparable to Panda-70M, providing a similar order of magnitude. 
However, our dataset specifically focuses on videos with spatio-temporal variations, incorporating preliminary human filtering to enhance quality.


\subsection{Video Segmentation}

Videos from the Internet are often excessively long and do not consistently feature camera movement.
To address this, we develop an automatic extraction tool based on a heuristic method to efficiently detect usable segments.
Specifically, we design a camera movement detection program leveraging optical flow estimation between adjacent frames. 
The program identifies camera motion by measuring optical flow displacement and retains segments where displacement exceeds a predefined threshold. Continuous sequences of such frames are then extracted to isolate periods of camera movement.
To ensure sample continuity, we impose an upper limit on the Euclidean distance of optical flow between adjacent frames, preventing abrupt scene transitions and hard cuts. 
This tool is implemented by extending PySceneDetect~\cite{PySceneDetect}.
Using this tool, we extract \textbf{107.6 million} video clips from \textbf{2.81 million} raw videos, averaging \textbf{38.29} clips per raw video.


\subsection{Video Clip Filtering}

\begin{figure}
    \centering
    \includegraphics[width=0.95\linewidth]{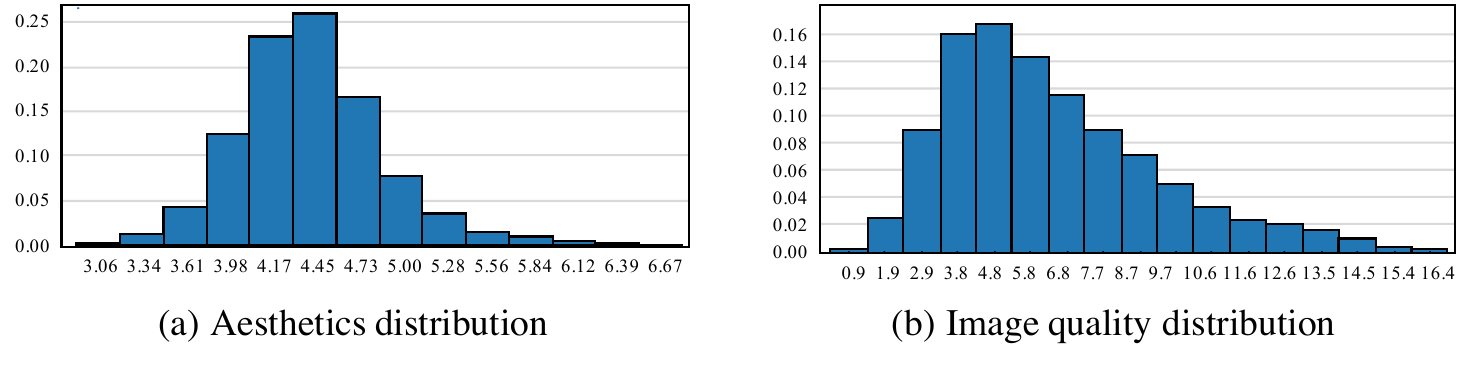}
    \caption{\textbf{The aesthetics distribution and the image quality distribution of \ourdataset{}.} These distributions demonstrate that our dataset achieves \textbf{high scores} in both aesthetics and image quality, indicating an overall \textbf{high-quality standard} for the dataset.}
    \label{fig:dataset_distribution}
\end{figure}

To facilitate video generation model training, we need to select high-quality spatio-temporal video clips from the automatically segmented videos.
Therefore, we developed a novel classification model, which classifies camera motion types based on the observed motion magnitude and style.
We define four primary categories: (C1) camera orbiting or target self-rotation, (C2) local horizontal or vertical tilting, (C3) camera tracking a moving target, and (C4) linear camera motion.
Additionally, we classify clips with static or near-static camera movement as (C5) and those edited using software, such as transitions or artificial effects, as (C6). 
To ensure high-quality data, we exclude most C5 clips and all C6 clips from \ourdataset{}.

To automate this process, we manually label 20,000 video clips and train a classification model based on the Video Swin Transformer\cite{liu2022swin}.
We use this model to identify and categorize clips belonging to the four primary motion types, forming a spatio-temporal-aware video dataset.
Additionally, we included a small proportion (less than 5\%) of aesthetically pleasing and high-quality videos from class C5, as these clips contribute to enhancing the overall quality of video generation.

Next, we refine the dataset by selecting high-quality videos based on aesthetics and image quality.
We utilize the publicly available LAION aesthetics model~\cite{schuhmann2022laion} to compute aesthetic scores and the DOVER-Technical model~\cite{wu2023exploring-dover} to evaluate image quality.
Only clips surpassing predefined thresholds are retained.
The distributions of aesthetics and image quality scores for \ourdataset{} are illustrated in Fig.~\ref{fig:dataset_distribution}.
Notably, nearly 95\% of clips achieve an aesthetic score above 3.5, while approximately 78\% exceed a score of 4.0 in image quality, underscoring the dataset's high visual fidelity.

\subsection{Video Captioning}

\begin{figure}
    \centering
    \includegraphics[width=1.0\linewidth]{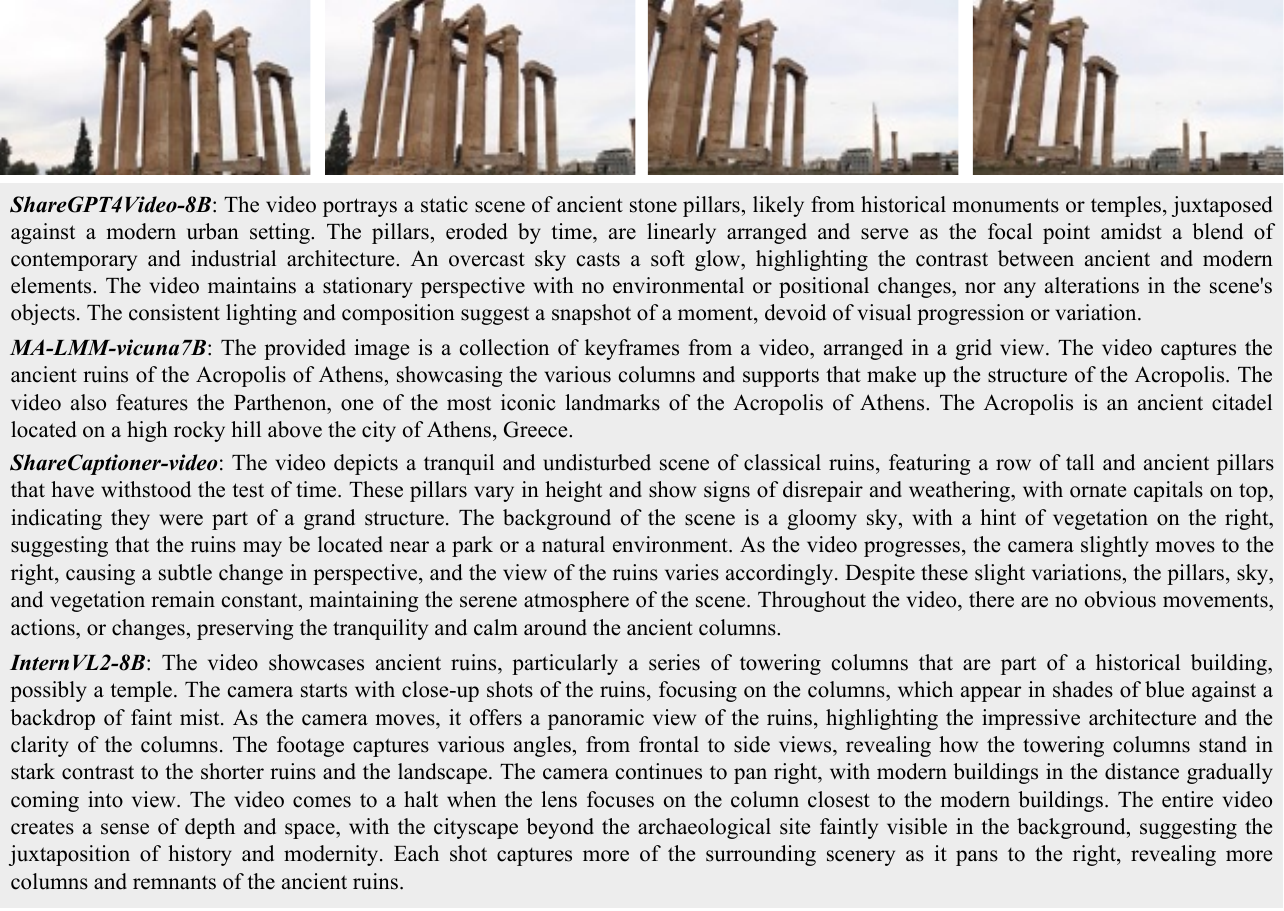}
    \caption{\textbf{Captions generated by the fine-tuned models,} including InternVL2-8B\cite{chen2024far,chen2024internvl}, ShareGPT4Video-8B\cite{chen2024sharegpt4video}, ShareCaptioner-video\cite{chen2024sharegpt4video}, and MA-LMM\cite{he2024ma}. InternVL2-8B\cite{chen2024far,chen2024internvl} captures intricate camera work and narrative elements with high efficacy.}
    \label{fig_caption_compare}
\end{figure}

\begin{figure}
    \centering
    \includegraphics[width=0.99\linewidth]{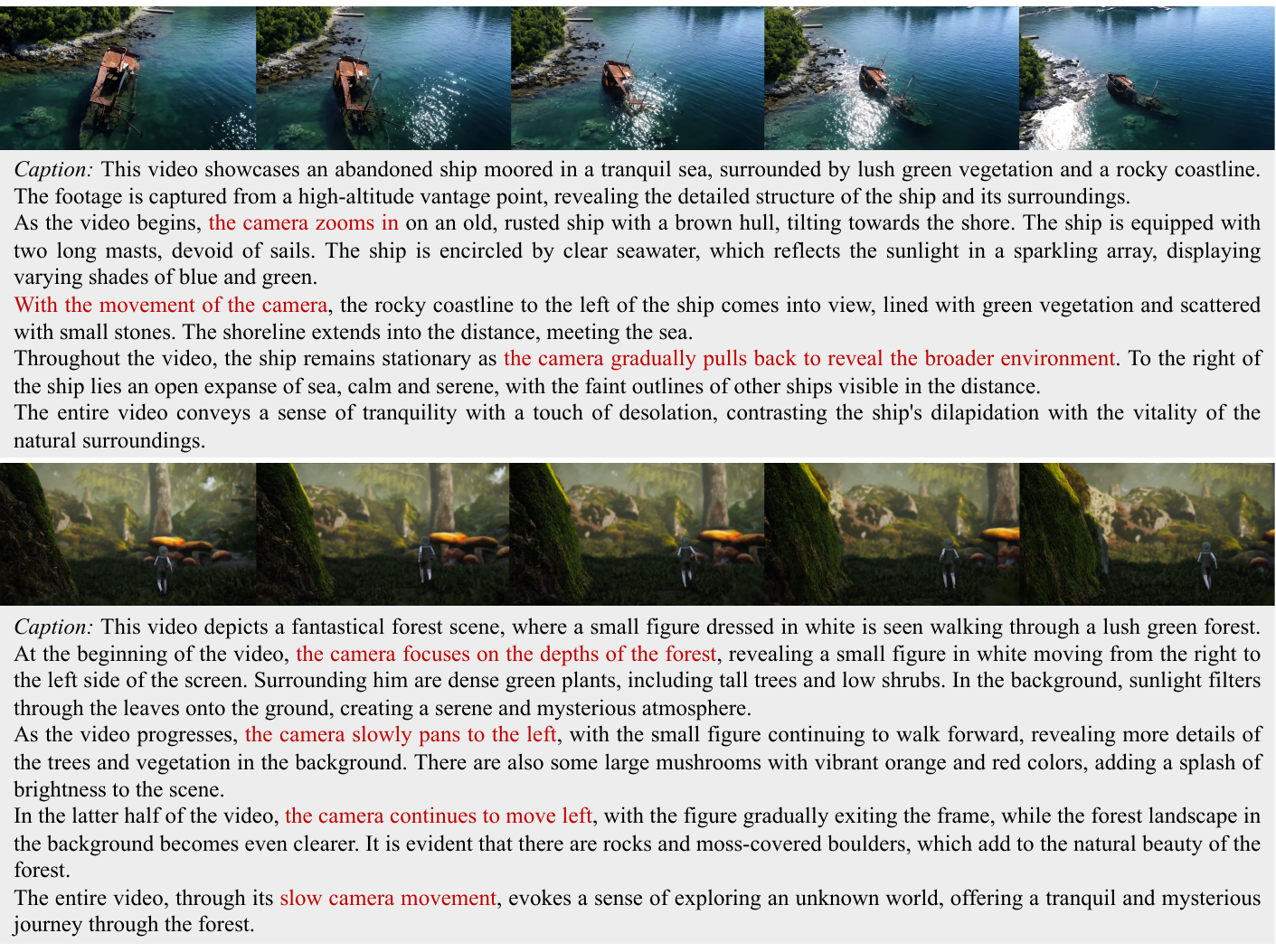}
    \caption{\textbf{Results of the fine-tuned video captioning model.} In the prompts, descriptions related to camera motions are highlighted in \textcolor{red}{red}. It is evident from the training samples that the camera undergoes multiple motion changes. Moreover, the scene details in the videos are clearly described and accurately followed as the camera moves. These high-density informational text captions significantly enhance the spatio-temporal semantics of the videos. Consequently, our video captions in the \ourdataset{} dataset provide enriched guidance for training video generation models.}
    \label{fig_dataset_caption}
\end{figure}

We employ a video-to-text model to generate captions for video clips, reducing the need for extensive human labor.
However, existing video-to-text models typically produce brief descriptions, which are insufficient for ensuring spatio-temporal consistency in our video generation task.
To address this, we first curate a dataset of videos with captions that provide comprehensive descriptions of objects, scenes, and visual transitions, with a particular emphasis on camera movements and their effects.
Subsequently, we utilize GPT-4 to correct grammatical and spelling errors in the captions, thereby creating a dataset suitable for instructionally supervised fine-tuning of multimodal models.
Based on this dataset, we fine-tune several open-source multimodal models, each containing approximately 7–9 billion parameters, including InternVL2-8B\cite{chen2024far,chen2024internvl}, ShareGPT4Video-8B\cite{chen2024sharegpt4video}, ShareCaptioner-Video\cite{chen2024sharegpt4video}, and MA-LMM\cite{he2024ma}.
Fig. \ref{fig_caption_compare} illustrates the outcomes of these models.

We evaluate these models using human expert ratings.
Both InternVL2-8B and ShareCaptioner demonstrate strong performance, excelling at generating detailed and precise descriptions of camera movements while maintaining semantic richness and coherence.
However, ShareCaptioner-Video exhibits significantly reduced efficiency due to its sliding captioning and clip summarization strategy, which requires distinct descriptions for each sampled frame, leading to more frequent LLM invocations.
Balancing efficiency and performance, we selected the fine-tuned InternVL2-8B for large-scale caption generation in the \ourdataset{} dataset.
This improved model generates highly detailed descriptions that accurately capture interactions caused by lens changes, including camera movements, various transitions, and content shifts, thereby providing precise training data for video generation models.

Fig. \ref{fig_dataset_caption} presents two complete samples, illustrating that the generated captions comprehensively describe camera operations and the visual transitions induced by motion. Additionally, we ensured that descriptions include sufficient details about lighting, style, and atmosphere of objects and backgrounds, thereby offering richer guidance for model training.
Each video segment is annotated with captions averaging 206 words in length, ensuring a high level of detail and descriptive accuracy.

\section{The DropletVideo Model}

\begin{figure}[htbp]
\centering
\includegraphics[scale=0.6]{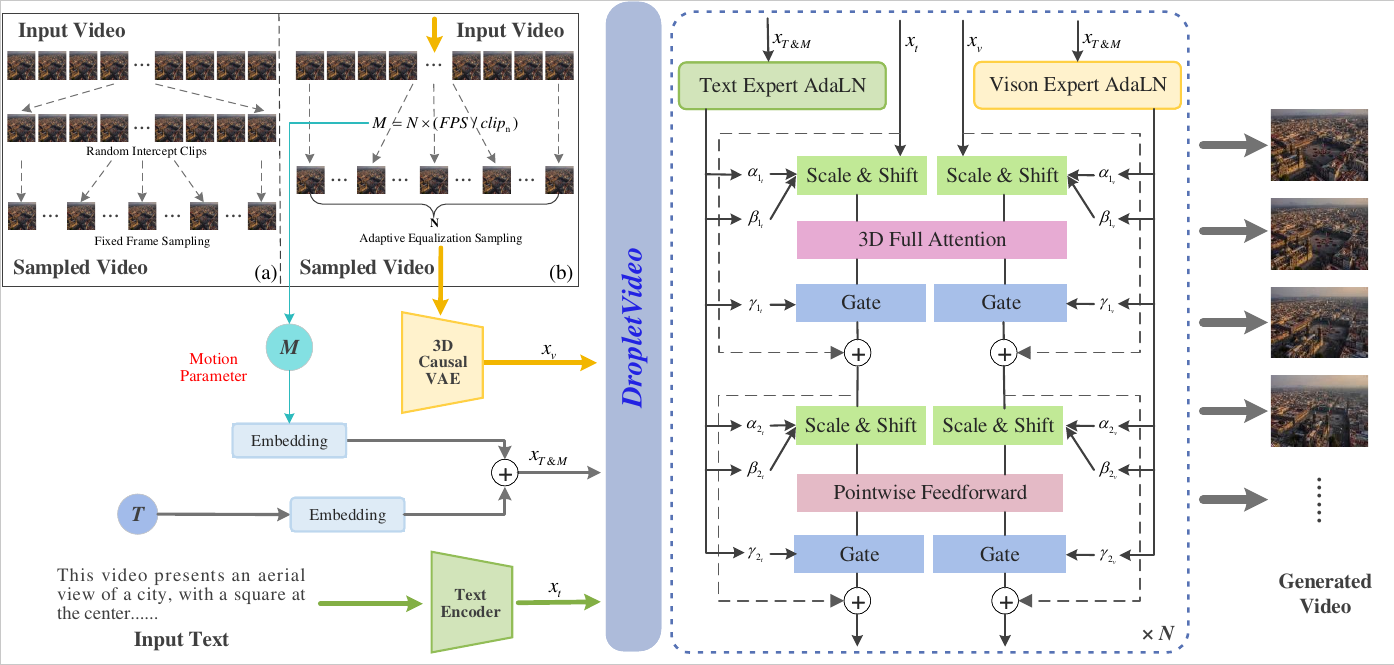}
\caption{\textbf{Overview of the \ourproj{} Framework.} 
The video is processed by the 3D causal Variational Autoencoder (VAE) following adaptive equalization sampling, which is steered by the motion intensity $M$. 
The video feature $x_v$ is then input into the Modality-Expert Transformer, depicted on the right side of the figure, to facilitate video generation in conjunction with the text encoding $x_t$, and the combined encoding $x_{T\&M}$ of the temporal $T$ and the motion intensity $M$. 
The upper left part illustrates the contrast between (a) \textbf{the traditional sampling approach} and (b) \textbf{\ourproj{}'s adaptive equalization sampling}. 
Traditional methods involve random segment interception followed by fixed-frame-rate sampling of the intercepted segments, whereas \ourproj{} employs adaptive frame rate sampling across the entire video segments, guided by $M$.}
\label{figure:DropletVideoframework}
\end{figure}

Using the videos and captions from \ourdataset{}, we train a video generation model designed to preserve both temporal and spatial consistency, with a particular emphasis on camera angles and object movement.
We name this model \ourproj{}, and its architecture is illustrated in Figure~\ref{figure:DropletVideoframework}.


\subsection{Preliminary Overview of Diffusion Models}
The proposed \ourproj{} is developed and trained utilizing diffusion model (DM)\cite{dhariwal2021diffusion}.
The essence of a DM involves generating samples from a distribution by reversing a gradual noising process. 
This process initiates with a noisy input, \( x_T \), which is usually Gaussian noise, and sequentially produces less noisy samples, \( x_{T-1}, x_{T-2}, \dots \), culminating in the final sample, \( x_0 \). 
The timestep \( t \) is used to indicate the noise level. \( x_t \) represents a combination of the original signal \( x_0 \) and added noise \( \epsilon \).


During the diffusion phase, the model progressively adds noise to the data, increasing in intensity until the original data is fully transformed into Gaussian noise.
Given a real data distribution $x_0$$\sim$$q(x)$, and it is sampled $T$ times to add Gaussian noise. 
The variation schedule of the noise is defined as $a_t$, and the data thus sampled is denoted as $x_t$, where $t\in[1,T]$. 
The process obeys a Markov chain, and after a reparameterization trick, the model can directly obtain any intermediate state, and the sampling formula for $x_t$ is $q(x_t)=N(x_t;\sqrt{\bar{a_t}}x_0,(1-\sqrt{\bar{a_t}})I)$, where $\bar{a_t}=\mathop{\Pi}\limits_{i=1}^{t}a_i$.

Conversely, during the denoising phase, the model learns the real data distribution from the standard Gaussian noise \( p(x_T) \), where \( p(x_T) = \mathcal{N}(x_T; 0, I) \). 
The DM is trained to generate a successively denoised \( x_{t-1} \) from \( x_t \). 
Ho et al.\cite{ho2020denoising} define the model as a function \( \epsilon_\theta(x_t, t) \) that estimates the noise component in the noisy sample \( x_t \). 
The noise prediction function \( \epsilon_\theta(x_t, t) \) is usually obtained by designing a U-Net network stacked with residual networks. 
The optimization objective is then defined as
$\|\epsilon_{\theta}(x_t, t) - \epsilon_t\|^2$
, where \( \epsilon_t \) represents the sampled noise at time \( t \) and serves as the ground truth.

To mitigate the high computational and resource demands of conventional diffusion models in generating high-dimensional data, a series of latent diffusion models (LDMs)\cite{rombach2022high} has been introduced. 
A LDM employs a pre-trained perceptual compression model consisting of an encoder $\varepsilon$ and a decoder $D$\cite{preechakul2022diffusion,bengesi2024advancements}. 
This integration allows the diffusion process to transfer from the high-dimensional pixel space to the low-dimensional latent space, thereby enabling learning in the latent representation domain.
The objective function of the LDM is $L_{LDM}=E_{\varepsilon(x_0),t,\epsilon_{\theta}\sim{N(0,I)}}[\Vert\epsilon_t-\epsilon_\theta(z_t,t)\Vert^2]$, where $z_t$ is the output of the encoder.

Drawing inspiration from 3D Variational Autoencoders\cite{yu2023language}, \ourproj{} model encodes video frames into the latent space using three-dimensional convolutions, capturing both spatial and temporal dimensions. 
Additionally, we incorporate the Multi-Modal Diffusion Transformer (MMDiT) model\cite{esser2403scaling}. 
This integration permits the model to function autonomously within the representation spaces of text and video, while also accounting for their inter-dependencies, thereby facilitating enhanced information transfer and synthesis.

\subsection{Architecture}
The architecture of \ourproj{} is shown in Fig. \ref{figure:DropletVideoframework}.
During the training process, the input consists of a textual prompt, a video, a time parameter $T$, and a motion-control parameter $M$. 
The multi-modal inputs are embedded into the latent feature space through the corresponding encoders, respectively. 
The text encoder is T5\cite{raffel2020exploring} and the 3D causal VAE is applied for visual information. 
Subsequently, the potential features, $T$, and $M$ are embedded into the modality-expert transformer architecture, respectively. 
Finally, a new video, satisfying the desired motion speed, is decoded by the denoised latents, with a 3D causal VAE decoder. 

\subsubsection{3D Causal VAE}
Different from other auto-encoders, the outputs of VAE's encoder and decoder are subject to parameter-constrained probability density distributions. 
Yang et al.\cite{yang2024cogvideox} applied 3D convolutions to video reconstruction. It is demonstrated that the 3D structure can reduce the jitter problem in the reconstructed video. 
Therefore, in \ourproj{} architecture, we apply 3D causal VAE extended with VAE and 3D structure to the encoding and decoding of video frames. 
It reduces the computation of \ourproj{} and ensures the efficiency and continuity of the generated video.

\subsubsection{3D Modality-Expert Transformer}
The input for \ourproj{} consists of two modalities, textual prompt and video. 
To ensure smooth embedding of each modality, 3D positional embedding is applied in the transformer architecture, and multi-modal attention is employed to handle text and vision data simultaneously. 
3D full attention is a technique that has evolved with the widespread application of transformer in computer vision, and we apply it to \ourproj{}. 
Compared with the previous separation approach, it can better capture dynamic variations in the video and enhance the semantic consistency and diversity of the generated content.

\subsubsection{Motion Adaptive Generation}
To address the challenge of generating videos with varying motion speeds, we innovatively introduce the Motion Adaptive Generation (MAG) strategy within our \ourproj{} model. 
This strategy allows the model to dynamically adjust to the desired speed of motion in the generated video content.

The generated videos from previous models usually have a fixed motion speed, mainly because they adopt a fixed frame rate to sample the raw video frame, as shown in Fig. \ref{figure:DropletVideoframework} (a). 
Previous models first sample a sub-clip from the original video-clip and then sample the video frame according to fixed FPS (for example, selecting one frame every three frames). 
However, it fails to meet the customers' requirements for more details presented on the video. 
To generate more visually appealing videos, MAG is designed in \ourproj{} to ensure that the generated video is motion-controlled. 
Here, we uniformly sample video frames over the entire video stream and adopts the detailed caption data for these sampled frames, thus capturing global dependencies and obtaining more complete semantic information, as shown in Fig. \ref{figure:DropletVideoframework} (b). 
We introduce the motion intensity \( M \), which is used to control the motion intensity of the generated video. \( M \) can be defined as follows

\[
M = N \times \left( \frac{FPS}{clip_n} \right),
\]

where \( FPS\) represents the FPS of the video, \( clip_n \) represents the number of video frames, N represents the sample number during the training process.

In the \ourproj{} framework, the MAG strategy jointly modulates the input coding with time $T$. 
Since the feature states of the two input modalities, text and videos, are quite diverse, we apply the text expert adaptive layernorm (Text Expert AdaLN) and vision expert adaptive layernorm (Vison Expert AdaLN) strategies independently in the text and vision latent spaces. 
Through the auxiliary calculation of these two modules, the control parameter $M$ is smoothly delivered to the diffusion transformer, and \ourproj{} can precisely control the dynamics of the generated video.

\section{Experiments}

\begin{figure}[h]
    \centering
    \includegraphics[width=1.0\linewidth]{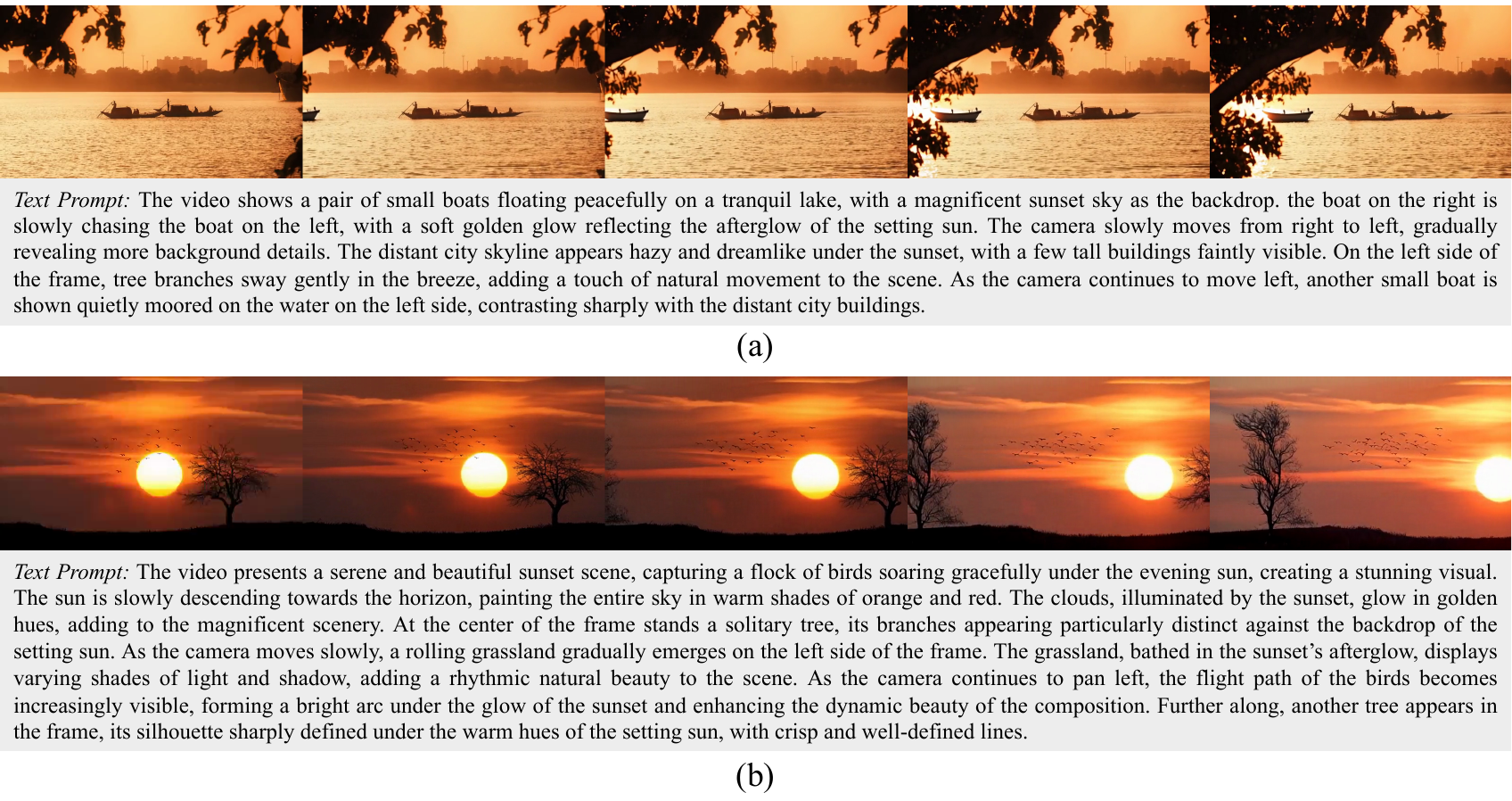}
    \caption{\textbf{\ourproj{} facilitates the generation of videos that maintain integral spatio-temporal consistency.} New objects or scenes introduced via camera movement are seamlessly integrated and interact logically with the pre-existing scenes. In video (a), as the camera moves, a new boat appears on the lake, the boat on the right of the original two boats continues to slowly chase the boat on the left, and the leaves on the shore still sway gently in the breeze. In video (b), as the camera moves left, the tree called for in the text prompt successfully appears in the shot, the original flock of birds continues to fly, and the grass and sky show continuity as the camera moves.}
    \label{fig:integralST}
\end{figure}

\subsection{Implementation Details}

During training, videos were resized according to different token lengths. In the first phase, the maximum token length was set to 13,312, supporting the generation of 49 video frames at a spatial resolution of 512 $\times$ 512. 
In the second phase, the maximum token length was increased to 68,992, enabling the generation of 85 frames at 896 $\times$ 896 resolution. 
To mitigate transition effects from initial frames, the first and last 10\% of frames were trimmed before uniform sampling.

We adopted the pre-trained \texttt{CogVideoX-Fun}\cite{cogvideox-fun} model for weight initialization.
Besides, we employed \texttt{t5-v1\_1-xxl}\cite{raffel2020exploring} as the text encoder. 
The maximum text tokenizer length was set to 400 instead of 226 to accommodate longer captions. 
The model architecture was based on the MMDiT series\cite{esser2403scaling}, consisting of 42 layers with 48 attention heads, each with a dimension of 64. 
The time step embedding dimension was set to 512. 
For optimization, we used \texttt{Adam}\cite{kingma2014adam} with a weight decay of 3e-2 and an epsilon of 1e-10. The learning rate was set to 2e-5. The number of sample frames (N) was fixed at 85. 
Training utilized the bfloat16 mixed-precision method with the DeepSpeed\cite{rasley2020deepspeed} framework. 
During inference, the classifier-free guidance scale was set to 6.5 to enhance temporal consistency and motion smoothness in the generated videos.

\subsection{Qualitative Evaluation}

\subsubsection{Integral Spatio-temporal Consistency}

\begin{figure}[h!]
    \centering
    \includegraphics[width=1.0\linewidth]{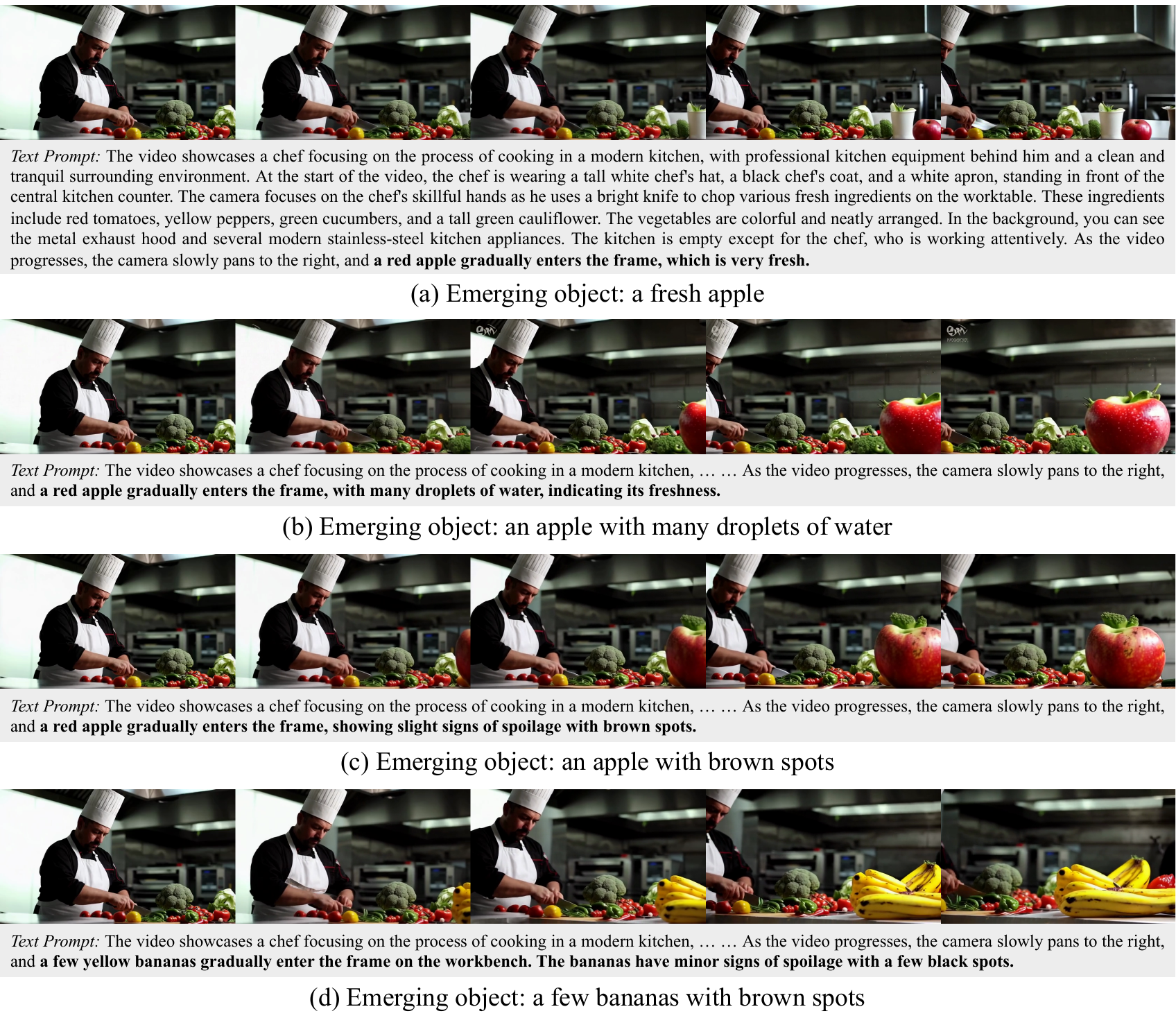}
    \caption{\textbf{\ourproj{} demonstrates advanced controllability in generating scenes where new objects emerges due to camera movement.} In video (a), as the camera pans right, the red apple specified in the prompt appears seamlessly, while the chef continues cooking, illustrating smooth integration of new objects. Video (b) showcases the system's ability to handle detailed descriptions, as the prompt's depiction of an apple with water droplets is rendered accurately, highlighting complex textures. In video (c), a prompt modification adds brown spots to the apple, which are visibly integrated, showing dynamic visual adjustments. Finally, in video (d), the prompt changes the apple to bananas, and the system adeptly features bananas, demonstrating versatility and precision in object transformation.}
    \label{fig:integralST_extend}
\end{figure}

\noindent{\textbf{Dynamic Scene Generation with Integral Spatio-temporal Consistency.}}
\ourproj{} focuses on integral spatio-temporal consistency during video generation. 
It addresses the spatial distortion issues caused by camera movement, ensuring smooth plot progression during camera movement and the spatio-temporal consistency of objects within the scene.
More importantly, in the development of a video scenario, the emerging scenes do not affect the behavior of the original video objects.
Fig. \ref{fig:integralST} exemplifies the integral spatio-temporal consistency. 
It is evident that \ourproj{} can maintain the continuity of the original plot while new plots enter the video.

\noindent{\textbf{High controllability of Emerging objects.}}
To further validate \ourproj{}'s capability in generating videos with integral spatio-temporal consistency, we conducted ablation studies focusing on the driving prompts. By modifying only the final sentences of the prompts while keeping the rest unchanged, we assessed the system's precision in controlling the characteristics of emerging objects, as shown in Fig. \ref{fig:integralST_extend}. The resulting videos clearly demonstrate \ourproj{}'s exceptional ability to accurately translate textual descriptions into visual elements, ensuring a high degree of fidelity to the specified attributes. This highlights \ourproj{}'s remarkable control over the emergence and detailed features of objects within the generated videos.

\subsubsection{3D Consistency}

\begin{figure}[h]
    \centering
    \includegraphics[width=1.0\linewidth]{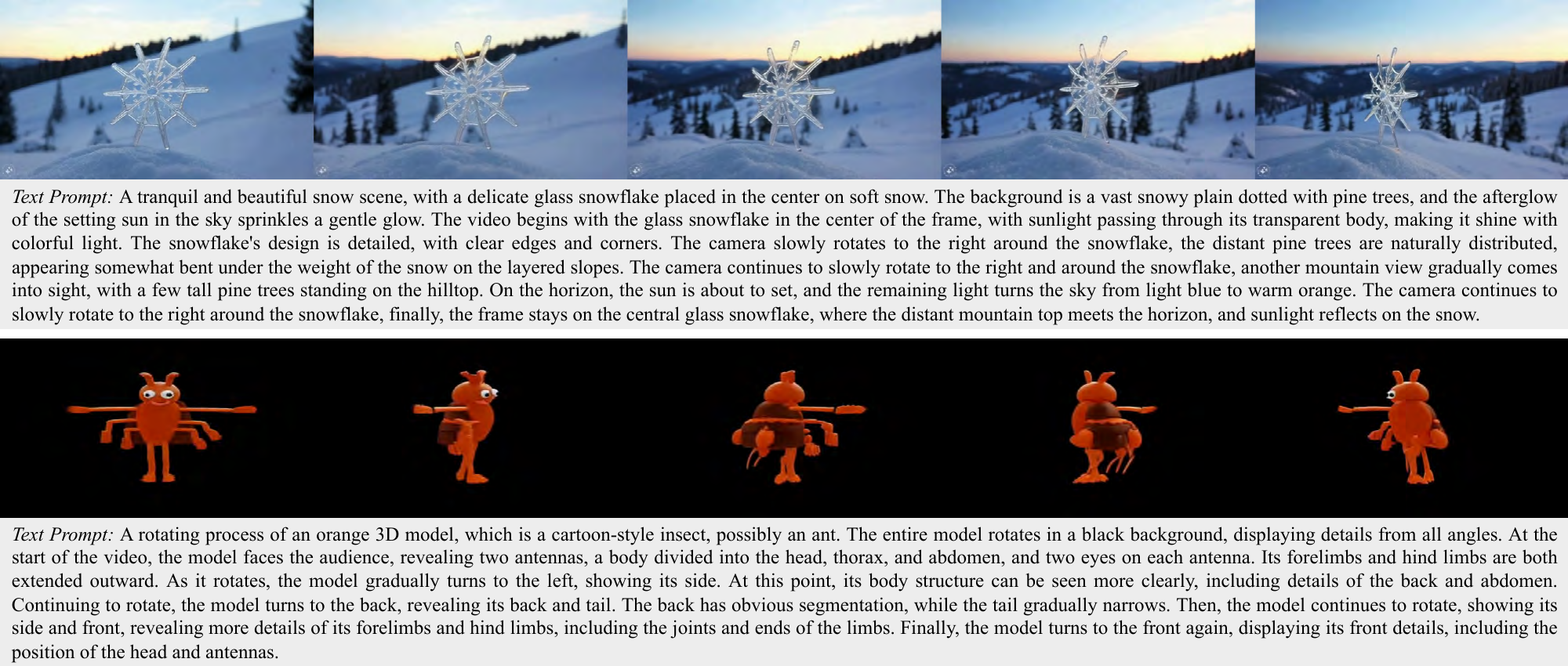}
    \caption{\ourproj{} demonstrates excellent 3D consistency. In the top example, the camera moves around a snowflake, showcasing significant camera movement while maintaining the snowflake's details from multiple perspectives. In the bottom example, the camera circles around an insect, and \ourproj{} ensures the insect's 3D consistency across a wide range of rotation angles. However, \ourproj{} still has limitations in generating content for a full 360-degree rotation, which will be addressed in future work. Overall, these examples illustrate \ourproj{}'s strong performance in spatial 3D consistency.}
    \label{fig:results_multiview}
\end{figure}

Trained on the large-scale spatio-temporal dataset, \ourdataset{}, \ourproj{} exhibits remarkable 3D consistency, as illustrated in Fig. \ref{fig:results_multiview}. In the top example, the camera rotates around a snowflake, maintaining stringent consistency for both the background and the snowflake from various angles, while preserving the snowflake's intricate details across multiple perspectives. In the bottom example, the camera performs an arc shot, projecting the same object. Despite not being specifically designed for arc shots, \ourproj{} effectively maintains the insect's 3D consistency over a broad range of rotation angles, demonstrating robust spatial 3D continuity.

\subsubsection{Controllable Motion Intensity}

\begin{figure}[h]
    \centering
    \includegraphics[width=1.0\linewidth]{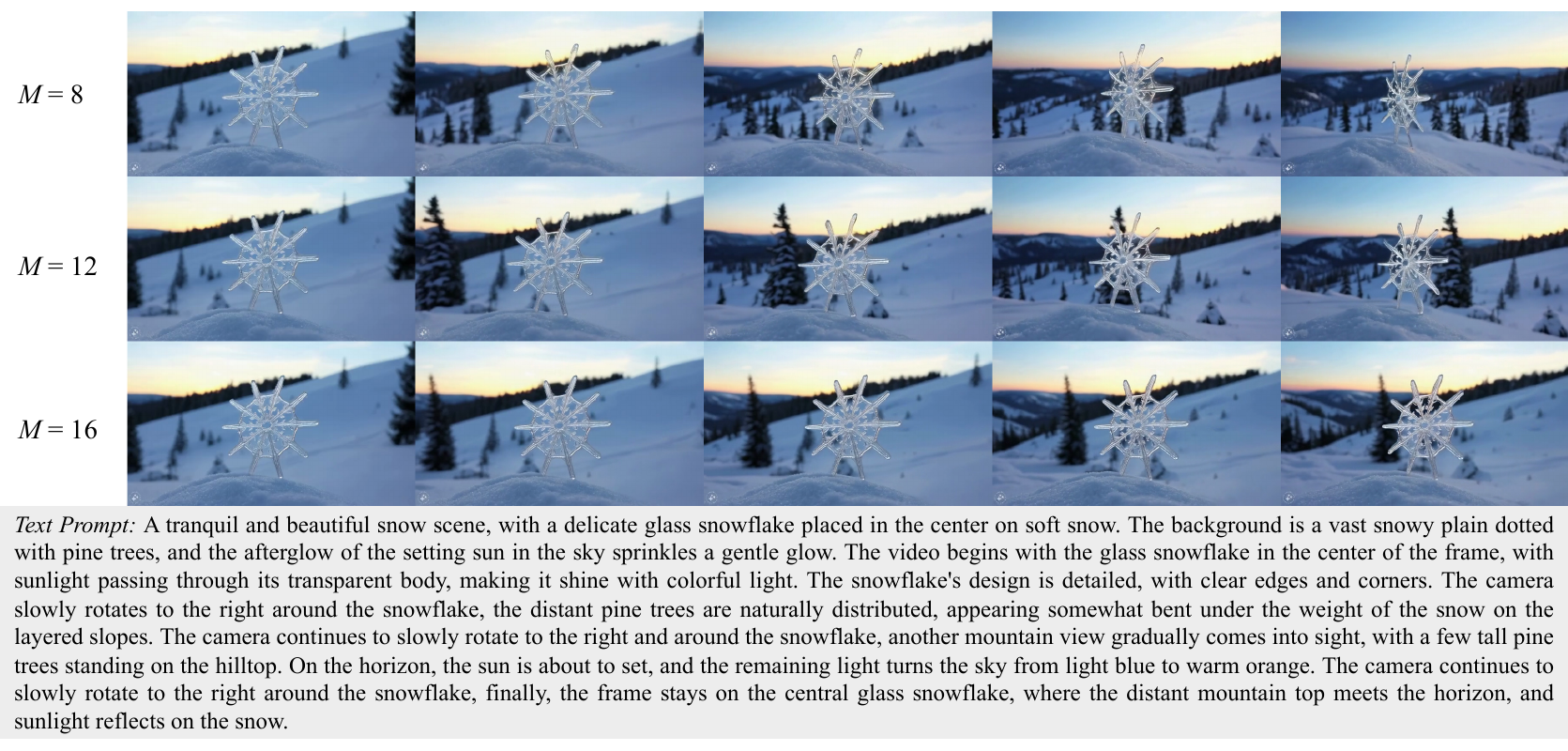}
    \caption{\ourproj{} facilitates precision control over video generation speed. Modifying the Input Speed parameter alters the movement speed of both the camera and target. In the third line, the camera motion parameter $M$ is doubled, and the snowflake's rotation speed is substantially decreased compared to the initial setting.}
    \label{fig:results_multifps}
\end{figure}

\ourproj{} manipulates the rate of plot progression and camera angle transitions through the adjustment of a motion control parameter. 
In the given example, enhancing this parameter allows a video of identical duration to accommodate more plot elements. 
Fig. \ref{fig:results_multifps} displays the video generation results under various motion control parameters using the same text-image input. 
Under the setting of $M=8$, the camera's movement is noticeably more pronounced than at $M=12$ and $M=16$, where the snowflake is presented with a broader range of perspectives. 
The motion density decreases as the $M$ escalates from 8 to 16, confirming that a lower $M$ results in a video with more drastic camera variations. 
This evidence suggests that \ourproj{} can adeptly regulate the playback speed of the content while maintaining semantic accuracy.

\subsubsection{Camera Motion}

\begin{figure}[h!]
    \centering
    \includegraphics[width=0.95\linewidth]{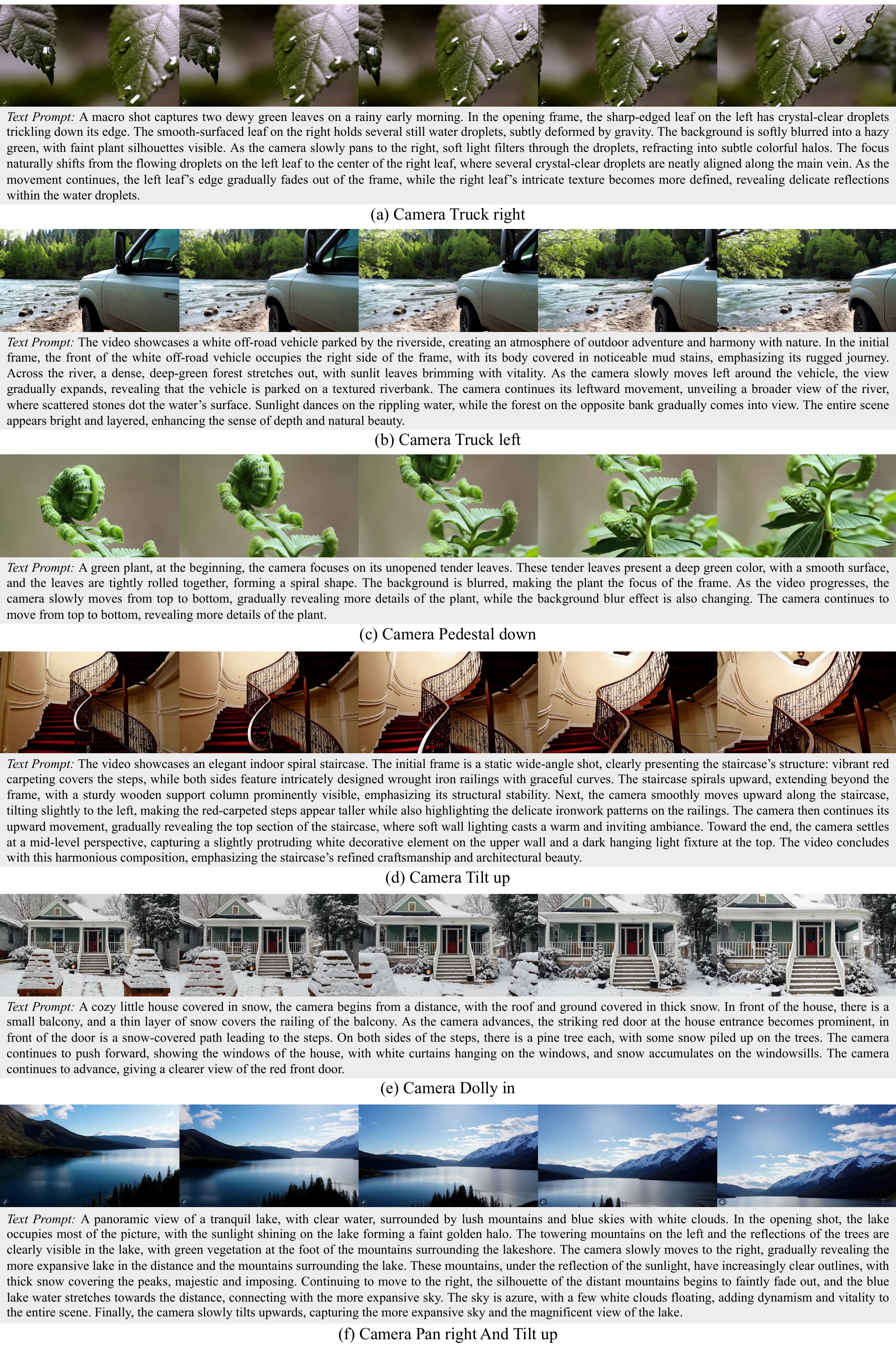}
    \caption{\textbf{\ourproj{} showcases its robust capabilities in generating videos with diverse camera movements.} Panels (a)-(e) illustrate the outcomes of specific camera motions: Camera Truck Right, Camera Truck Left, Camera Pedestal Down, Camera Tilt Up, and Camera Dolly In. Panel (f) presents a composite camera shot that combines Camera Pan Right and Tilt Up.}
    \label{fig:results_camera_shots_merge}
\end{figure}

\ourproj{} demonstrates versatile camera motion generation capabilities including various fundamental movement types, as visualized in Fig. \ref{fig:results_camera_shots_merge}. 
The system produces cinema-standard motions including right/left trucking, vertical pedestal movement, tilt adjustment, axial dollying, and composite pan-tilt operations.

\textbf{Camera truck right.} Fig.\ref{fig:results_camera_shots_merge} (a) illustrates precise truck right control through foliage dynamics. 
Beginning with a micro view of dual leaves, the system executes text-guided trucking movement where left leaf edges fade proportionally as right venation textures emerge. 
Focus transitions between flowing and static droplets maintain optical continuity, with refractive stability persisting despite background bokeh deformation that adheres to lens physics.

\textbf{Camera truck left.} The riverbank case in Fig. \ref{fig:results_camera_shots_merge} (b) showcases environmental expansion during leftward movement. Initial partial riverbank frames progressively incorporate complete stone formations and canopy structures, maintaining geometric coherence between existing and generated elements. 
Vehicle mud stains preserve spatial consistency while water ripples develop accurate motion parallax relative to camera speed. 
Dynamic light refraction on aquatic surfaces replicates real fluid behavior, particularly in water droplet translucency during splash events.

\textbf{Camera Pedestal down.} Vertical control in botanical close-ups in Fig. \ref{fig:results_camera_shots_merge} (c) manifests through synchronized plant revelation. 
Descending motion coordinates with stem texture emergence, where curled leaves gradually unfurl following botanical growth patterns. 
Background vegetation blur intensifies proportionally to focal plane descent, matching professional lens depth-of-field characteristics. 
Waxy surface highlights migrate smoothly across the leaves, preserving material authenticity during viewpoint transitions.

\textbf{Camera Tilt up.} The architectural validation in Fig.\ref{fig:results_camera_shots_merge} (d) confirms 3D spatial awareness during upward tilting. 
The spiral staircase geometry remains intact with stable railing spacing and curvature radii, while newly revealed decorative elements scale according to perspective principles. 
Color constancy persists across lighting variations, evidenced by consistent carpet saturation and wall temperature. 
Chandelier glow attenuation follows inverse-square law principles, with wall decorations maintaining physically accurate diffuse reflections.

\textbf{Camera Dolly in.} The snowscape progression in Fig. \ref{fig:results_camera_shots_merge} (e) demonstrates axial movement precision. 
Forward camera motion proportionally reveals architectural details: initially obscured red doors gradually restore surface textures under natural light decay, while window reflections adjust intensity with viewing distance. 
Pine trees maintain spatial reference integrity, their parallax displacements creating authentic depth gradients between foreground snow paths and background vegetation.

\textbf{Camera Pan right And Tilt up.} Composite motion control in Fig. \ref{fig:results_camera_shots_merge} (f) achieves seamless transition from lakeside panning to skyward tilting. Initial rightward movement preserves accurate spatial relationships between water glare and mountain reflections, with snow distribution transitioning naturally. During axis transition, lake area proportion decreases geometrically while emerging cloud formations maintain pattern continuity. Altitude-dependent lighting differentiation enhances realism, where high-altitude cloud translucency contrasts distinctly with low-altitude texture density.

\subsubsection{Comparison of our \ourproj{} with existing Models}

\begin{figure}[h!]
    \centering
    \includegraphics[width=0.95\linewidth]{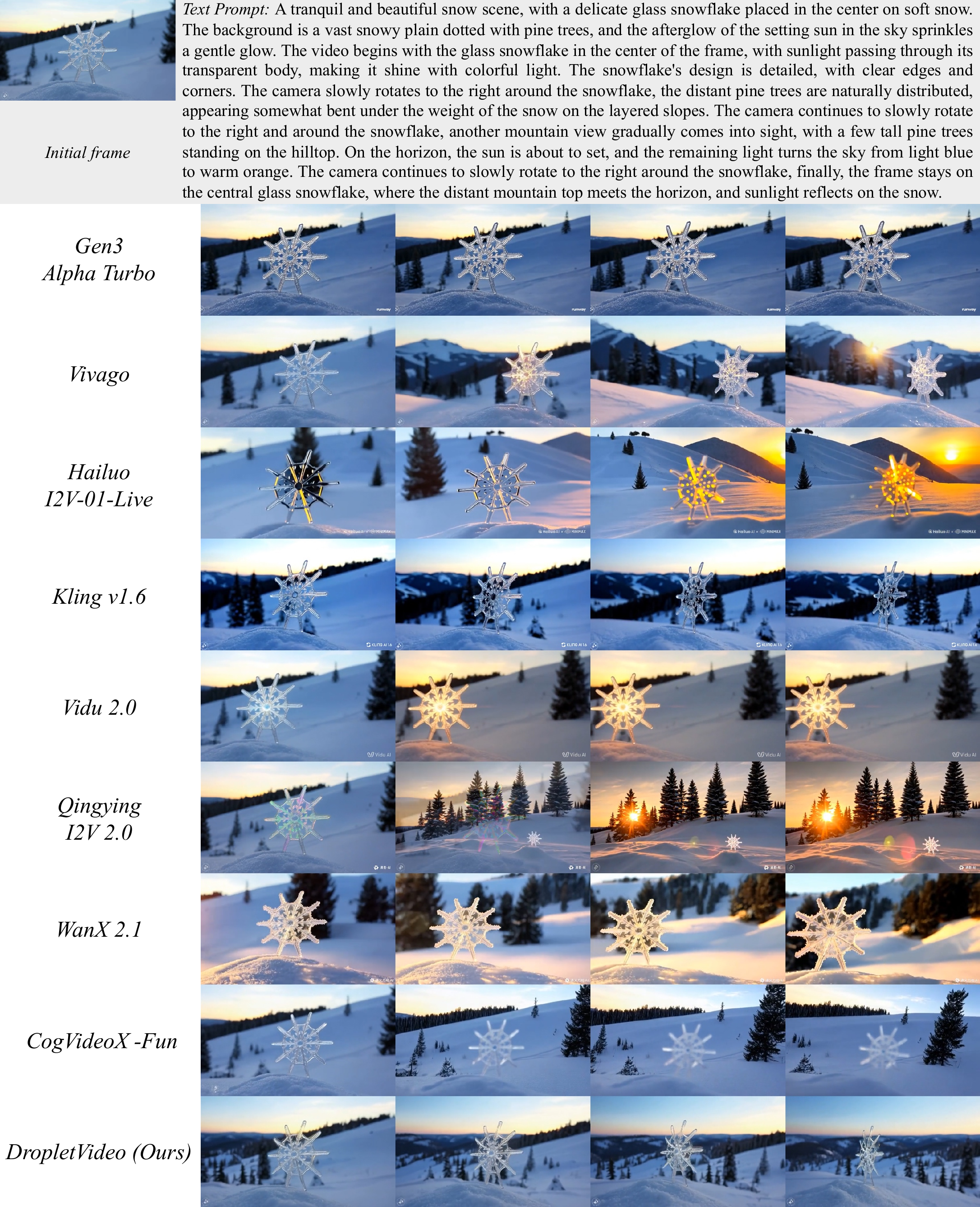}
    \caption{\textbf{Snow example.} The videos generated by \ourproj{}, Kling, and Vivago all maintain consistency with the prompt in terms of camera movement and various elements within the video. Their video quality is at the same level.}
    \label{fig:comparisons_merge_snow}
\end{figure}

\begin{figure}[h!]
    \centering
    \includegraphics[width=0.95\linewidth]{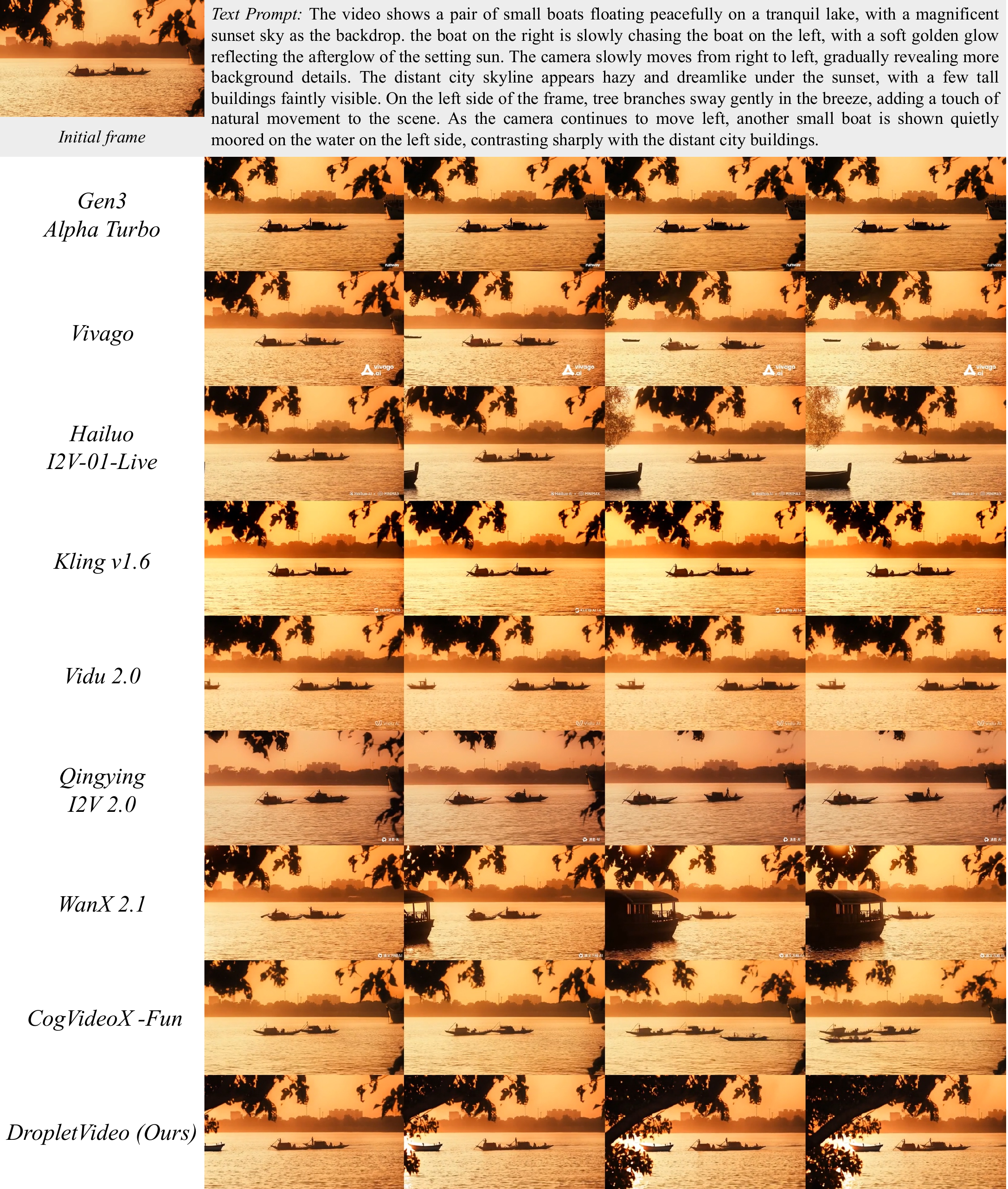}
    \caption{\textbf{Boat example.} Our \ourproj{}, along with Hailuo, WanX, and Kling v1.6, correctly understood the movement of the boat and the camera motion. However, these three models failed to ensure that the motion of the leaves remained logically consistent with the camera movement, resulting in the leaves moving synchronously with the camera, which is an unnatural effect. In contrast, our model maintains the relative motion consistency between the camera, boat, and leaves in the generated video. This is a typical demonstration of its integral spatio-temporal consistency capability.}
    \label{fig:comparisons_merge_boat}
\end{figure}

\begin{figure}[h!]
    \centering
    \includegraphics[width=0.95\linewidth]{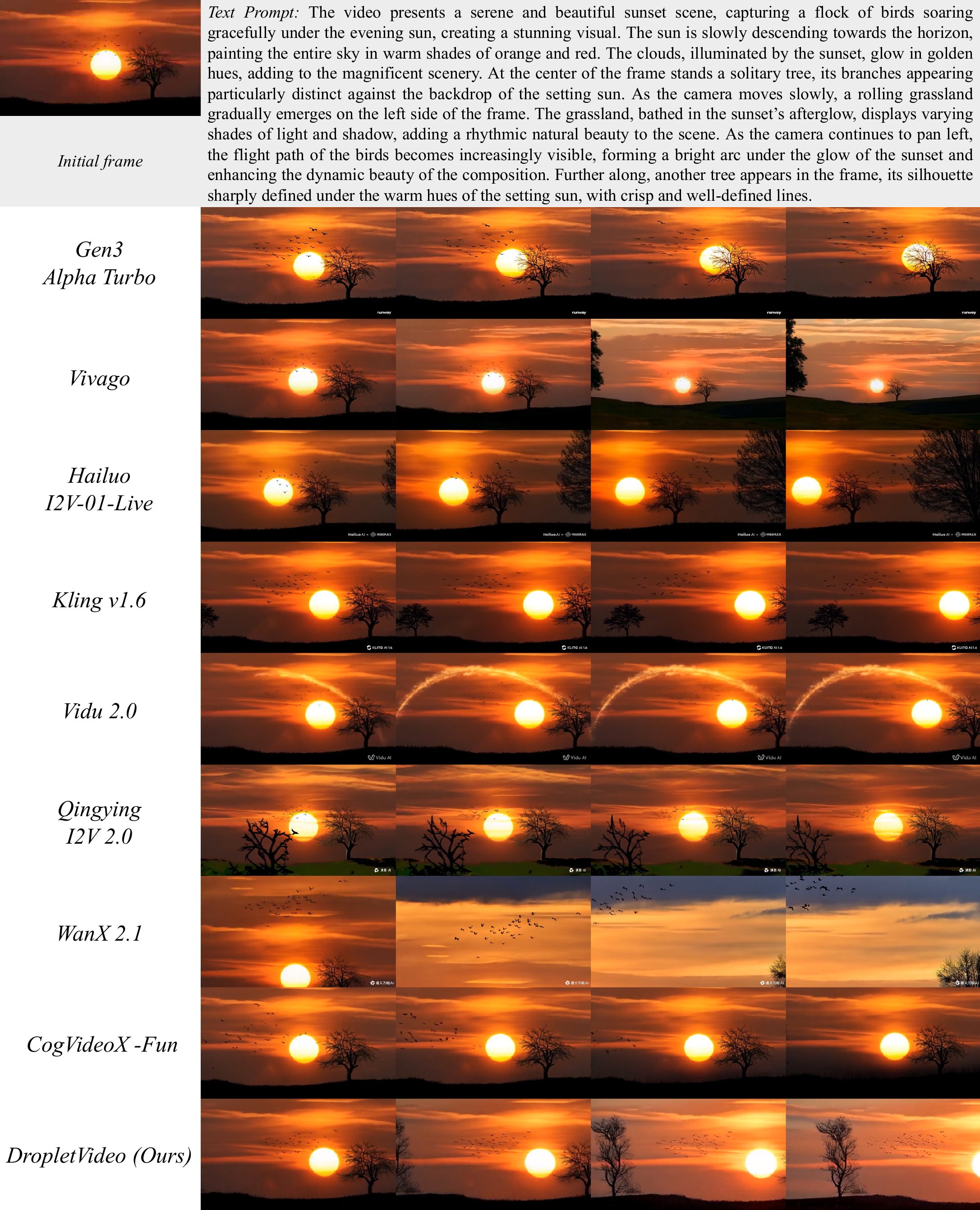}
    \caption{\textbf{Sunset example.} Only \ourproj{} and Kling v1.6 successfully ensure the correct alignment between camera movement and object positioning. However, in Kling's generated video, the lighting reflections on the clouds remain unchanged, lacking natural variation. In contrast, in our model's generated video, as the camera moves, the light reflections on the clouds dynamically adjust, making the scene more consistent with real-world natural phenomena.}
    \label{fig:comparisons_merge_sunset}
\end{figure}

\begin{figure}[h!]
    \centering
    \includegraphics[width=0.95\linewidth]{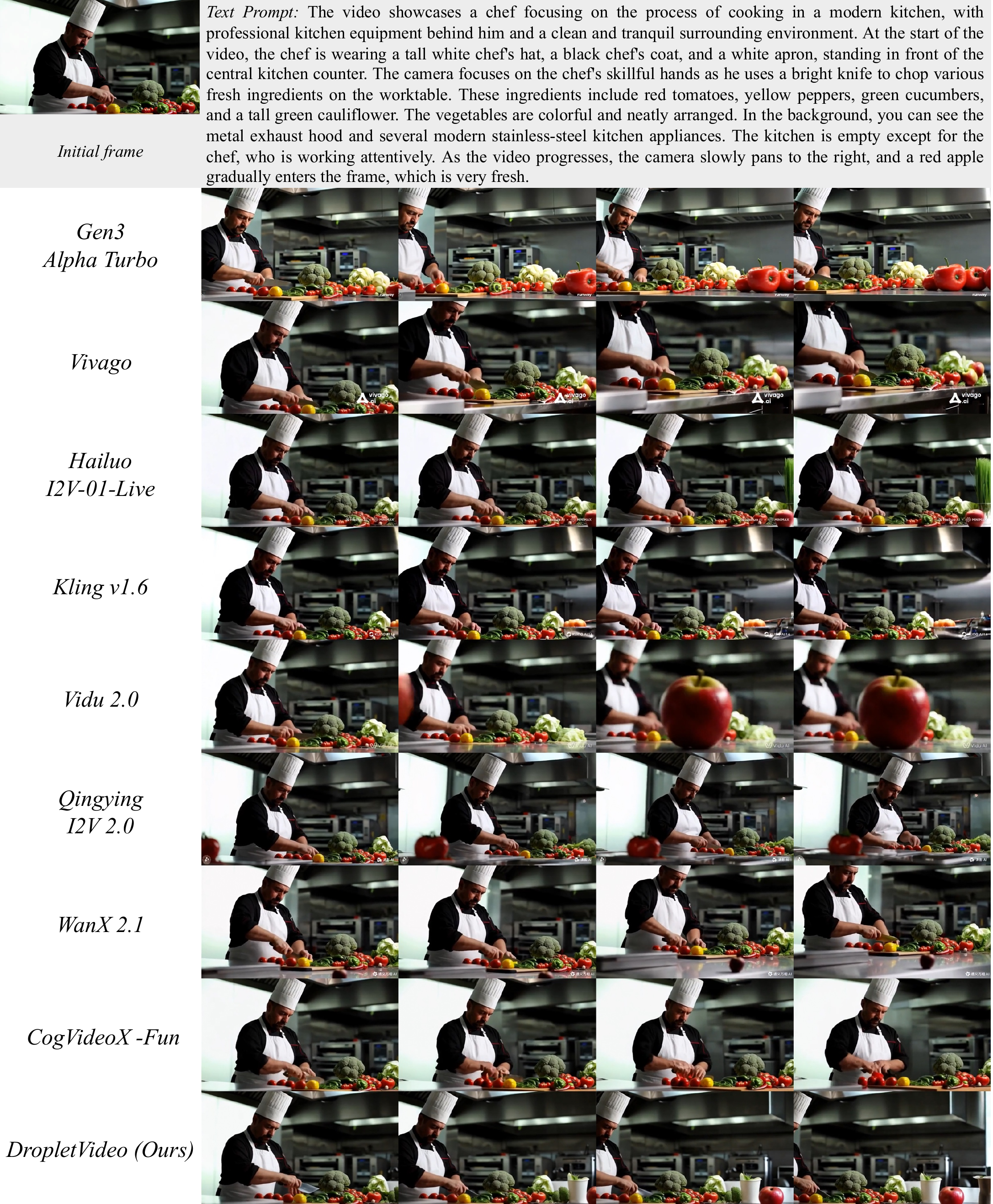}
    \caption{\textbf{Kitchen example.} We expect the focus of the video to transition from the chef to a red apple as the camera moves. Only \ourproj{} successfully achieved this transition, while other models failed to correctly generate ``a red apple'' after the camera movement. Besides, it also ensures that the apple it generates are of a reasonable size and are positioned appropriately within the scene.}
    \label{fig:comparisons_merge_kitchen}
\end{figure}

\begin{figure}[h!]
    \centering
    \includegraphics[width=0.95\linewidth]{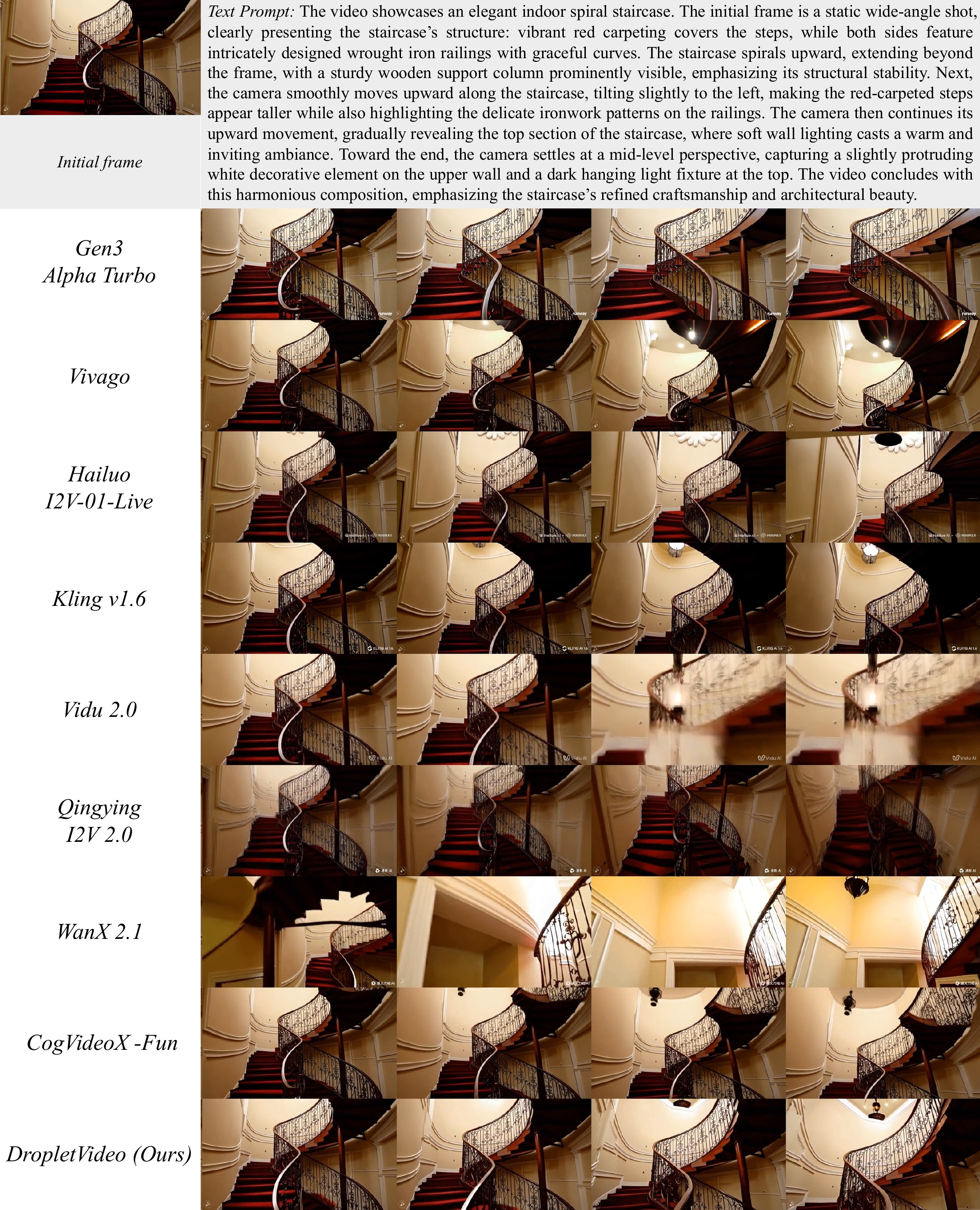}
    \caption{\textbf{Staircase example.} We required the camera to move smoothly up the stairs, ensuring that its trajectory remains logically consistent with the staircase in the video. Only our \ourproj{} and Gen3 successfully maintained the correct camera movement path. However, Runway failed to generate key elements such as wall decorations and lights.}
    \label{fig:comparisons_merge_staircase}
\end{figure}

\begin{figure}[h!]
    \centering
    \includegraphics[width=0.95\linewidth]{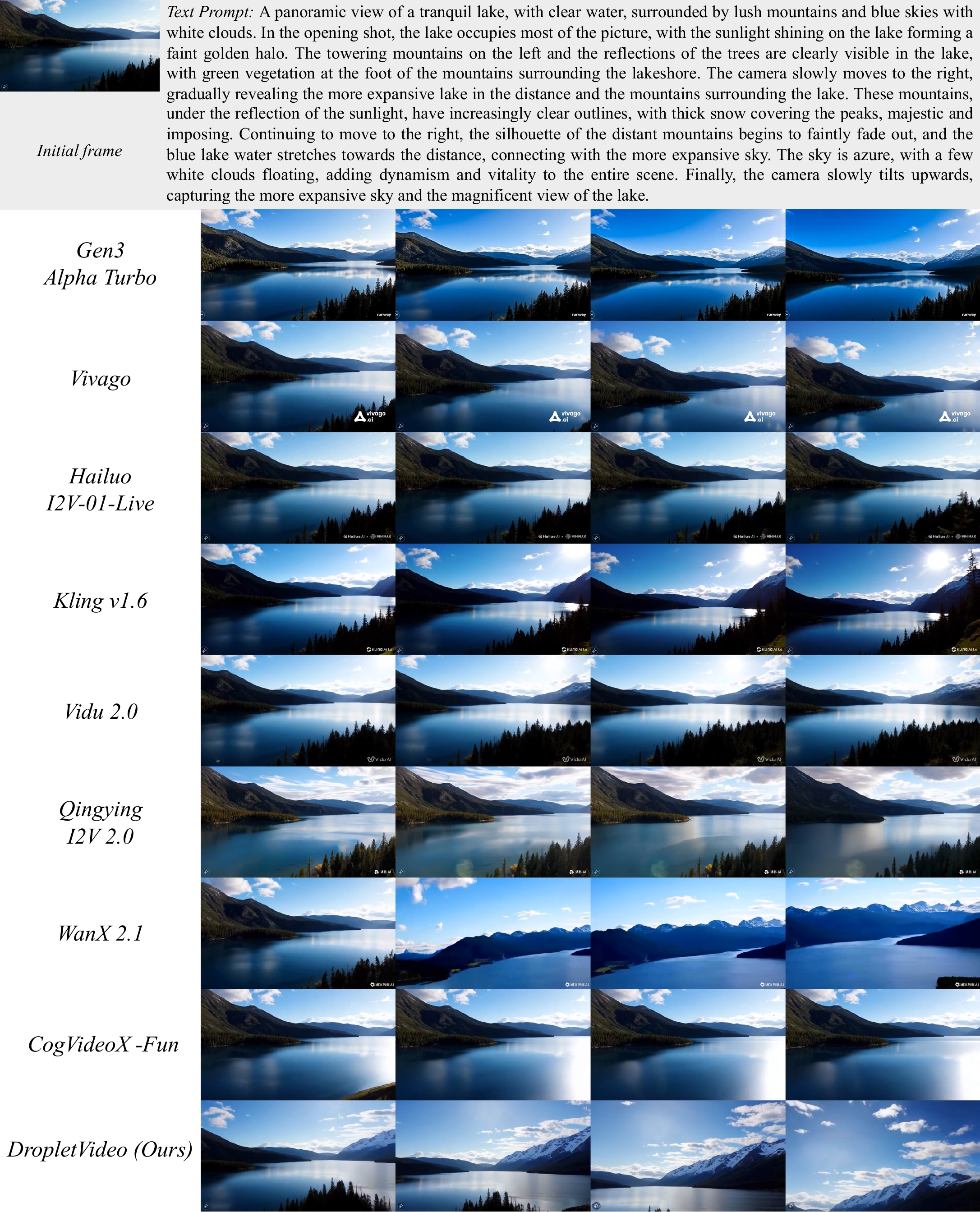}
    \caption{\textbf{Lake example.} The camera movement path is complex—it first moves to the right, then tilts upward, while the elements in the video change accordingly. All other models failed to accurately capture this camera movement, except for our \ourproj{}. Our model not only strictly followed the prompt in executing the camera motion but also dynamically altered the scene, successfully revealing the sky and white clouds, which were not present in the initial image.}
    \label{fig:comparisons_merge_lake}
\end{figure}

To better demonstrate the cumulative spatiotemporal consistency of \ourproj{}, we have selected several industry-recognized video generation models for comparison, including Hailuo \cite{hailuo},  Kling v1.6 \cite{kuaishou-klingai}, Gen-3 \cite{runway2024gen},  Vidu \cite{vidu}, Vivago \cite{vivago}, Qingying \cite{qingying}, CogVideoX-Fun \cite{cogvideox-fun}, and WanX \cite{wanx}.
Out of the compared models, only CogVideoX-Fun and WanX are open-source, similar to our approach, whereas the remaining models are closed-source.

We conducted comparisons using examples from various scenarios mentioned earlier, such as boat, kitchen, lake, snow, staircase, and sunset, as shown in Fig. \ref{fig:comparisons_merge_snow} - \ref{fig:comparisons_merge_lake}.

The examples of boat, sunset, and kitchen are particularly effective for evaluating cumulative spatiotemporal consistency, as they involve diverse spatial transformations and detailed descriptions of target features.
From these examples, we observe that WanX \cite{wanx} and Kling v1.6 \cite{kuaishou-klingai} perform relatively well. However, no single model excels across all these scenarios. In contrast, our algorithm consistently demonstrates superior spatio-temporal consistency across these examples.
For instance, as depicted in Fig. \ref{fig:comparisons_merge_boat}, \ourproj{} successfully produces a video where the camera rotation is precisely captured, simultaneously portraying the chasing interaction between the two boats. 
This level of detail and accuracy is beyond the capabilities of some other models, which struggle to generate such a scene with the same fidelity.

The scenarios of snow, staircase, and lake highlight \ourproj{}'s exceptional camera movement capabilities. 
Among the other algorithms, Kling v1.6 \cite{kuaishou-klingai} performs better, yet others fall short. 
Our algorithm, however, performs exceptionally well, closely adhering to the instructions given in the prompt.

Overall, despite \ourproj{} being an open-source model, it achieves and even surpasses the performance of existing well-known commercial generation models in terms of cumulative spatiotemporal consistency.
From this perspective, we believe that \ourproj{} holds greater promise in advancing the progress within the video generation community.

\subsection{Quantitative Evaluation}

\subsubsection{Dense Prompt Rewrite}\label{subsec_dpr}
To effectively address the variability in language style and length of user-provided prompts, and to offer detailed guidance for video generation, we implement a dense prompt generation preprocessing step.
This step serves as a bridge between the \ourproj{} system and user input.
Specifically, considering the superior performance of large language models in tasks such as text reasoning and image summarization, we have fine-tuned the InternVL2\cite{team2024internvl2} model with instruction tuning. 
This fine-tuning is done using the LoRA\cite{hu2021lora}, utilizing caption pairs from a high-quality training set. 
Experimental results indicate that approximately 600 such samples are sufficient to achieve the desired level of fine-tuning.

The module is designed to rephrase user prompts while keeping their original semantics intact. 
It transforms them into a standardized information architecture, akin to the trained captions. 
The module parses plot and camera movement details from the user input. 
It expands the content based on the input image, ensuring that the user's intent is preserved and detailed information is added. 
Furthermore, the module offers support for multiple languages.

We have revised 1,118 standard prompts supplied by VBench++~\cite{huang2024vbench++}, resulting in the same number of comprehensive prompts, which we have labeled VBench++-ISTP (Integral Spatio-Temporal Prompts). 
These revised prompts incorporate both temporal and spatial variations. 
For instance, consider the original VBench++ prompt: {\itshape``A couple of horses are running in the dirt.''} This has been rephrased to: {\itshape``The video showcases a dynamic scene of two horses running through mud, full of vitality and movement. The camera captures them kicking up dust, embodying a sense of freedom and abandon. The background faintly reveals the outlines of trees, adding a touch of natural tranquility to the entire scene. As the camera moves, the horses' running paths become clearer, and the dust sparkles in the sunlight, creating a dynamic visual effect.''} Compared to the original prompt, the rephrased version provides a more detailed depiction as the camera moves, effectively introducing spatio-temporal information.

\subsubsection{VBench++-IST Quantitative Results}

We carried out an extensive evaluation of \ourproj{}.
For this purpose, we employed the evaluation code and the core performance metrics supplied by VBench++\cite{huang2024vbench++}.
Furthermore, we integrated our integral spatio-temporal prompts, VBench++-ISTP, alongside the images from VBench++ \cite{huang2024vbench++}.
In particular, we have refined all prompts to include comprehensive detail, as mentioned in Sec. \ref{subsec_dpr}.
In our comparative analysis, \ourproj{} was benchmarked against the latest cutting-edge image-to-video models, including I2VGen-XL\cite{zhang2023i2vgen}, Animate-Anything\cite{lei2024animateanything}, and Nvidia-Cosmos\cite{agarwal2025cosmos}.

Quantitative results are presented in Tab. \ref{tab:table_quantitative_comparison}.
\ourproj{} outperforms the other three models in most performance metrics. 
In terms of I2V Subject, I2V Background and Motion Smoothness, \ourproj{}'s performance is 98.51\%, 96.74\%, and 98.94\% respectively, both surpassing the other three models. 
In the aspect of Camera Motion, \ourproj{} performs at 37.93\%, significantly higher than the 12.95\% of I2VGen-XL, the 10.64\% of Animate-Anything, and the 37.56\% of Nvidia-Cosmos.
This suggests a strong capability of \ourproj{} in handling camera motion within videos. 
For the Dynamic Degree, \ourproj{}'s performance surpasses I2VGen-XL and Animate-Anything, yet falls below Nvidia-Cosmos, indicating a competitive performance of \ourproj{} in maintaining motion coherence and dynamic degree.

\begin{table}[ht]
\centering
 \caption{\textbf{Comparison of \ourproj{} with state-of-the-art image-to-video models.} \ourproj{} outperforms other models in \textbf{I2V Subject}, \textbf{I2V Background}, \textbf{Motion Smoothness} and \textbf{Camera Motion}. Meanwhile, \ourproj{} remain at the current mainstream level for other metrics. In this table, \textbf{I2V-S} stands for I2V Subject, \textbf{I2V-B} stands for I2V Background, \textbf{CM} stands for Camera Motion, \textbf{SC} stands for Subject Consistency, \textbf{BC} stands for Background Consistency, \textbf{TF} stands for Temporal Flickering, \textbf{MS} stands for Motion Smoothness, \textbf{DD} stands for Dynamic Degree, \textbf{AQ} stands for Aesthetic Quality, \textbf{IQ} stands for Imaging Quality.}
\label{tab:table_quantitative_comparison}
\small
\begin{tabularx}{\textwidth}{l *{10}{>{\centering\arraybackslash}X}} 
\toprule
Models & \textbf{I2V-S} & \textbf{I2V-B} & \textbf{CM} & SC & BC & TF & \textbf{MS} & DD & AQ & IQ \\
\midrule
I2VGen-XL\cite{zhang2023i2vgen} & 96.08 & 94.67 & 12.95 & 95.76 & 97.67 & 97.40 & 98.27 & 24.80 & 65.26 & 69.21 \\
Animate-Anything\cite{lei2024animateanything} & 98.13 & 96.05 & 10.64 & 98.18 & 97.46 & 98.15 & 98.52 & 2.52 & 66.42 & 71.89 \\
Nvidia-Cosmos\cite{agarwal2025cosmos} & 95.10 & 95.30 & 37.56 & 91.59 & 94.43 & 96.20 & 98.82 & 83.90 & 58.39 & 70.35 \\
\midrule
\textbf{DropletVideo (Ours)} & \textbf{98.51} & \textbf{96.74} & \textbf{37.93} & 96.54 & 97.02 & 97.68 & \textbf{98.94} & 27.97 & 60.94 & 70.35 \\
\bottomrule
\end{tabularx}
\end{table}

In conclusion, despite some metrics where \ourproj{} falls short compared to other models, it exhibits significant advantages in most of the key performance metrics.
We believe that with further optimization and improvements, \ourproj{} will be able to reach or even surpass the performance of other advanced models.

\section{Conclusion}
In this paper, we introduce integral spatio-temporal consistency for the first time, which refers to the interaction between newly introduced objects due to camera movement and pre-existing ones.
We also released the largest spatio-temporal video dataset to date and open-sourced the foundational video generation model, \ourproj{}.
Experiments demonstrate that our approach exhibits a strong ability to achieve spatio-temporal consistency in video generation, surpassing most open-source models and even rivaling some closed-source models. 
Surprisingly, our model also exhibits a certain degree of 3D consistency. 

In future work, we will further investigate this issue by refining the data filtering strategies and expanding the dataset to a larger scale, with emphasis on diverse camera motions and dynamic objects.
Furthermore, the types of camera motions supported by VBench++\cite{huang2024vbench++} are very limited, which is insufficient to capture the richness of spatial variations.
It is worth exploring the development of a fine-grained camera motion classification model to better evaluate complex camera movements. Additionally, more suitable evaluation metrics should be proposed to comprehensively assess integral spatio-temporal consistency.
Additionally, given the model's strong 3D consistency capability, we plan to extend its application to 3D/4D content generation.



\bibliographystyle{plain}
\bibliography{sample}

\end{document}